\documentclass{article}

\usepackage{PRIMEarxiv}

\usepackage{amsmath,graphicx,amssymb}
\usepackage{hyperref}
\usepackage{pifont,cite}
\usepackage{caption}
\usepackage{booktabs}
\usepackage{xcolor}
\usepackage{subfigure}
\usepackage{tabularx}
\usepackage{subfigure}
\usepackage{algorithm}
\usepackage[noend]{algorithmic}
\def \etal {\textit{et al. }}

\usepackage[utf8]{inputenc}
\usepackage{graphicx}
\title{Versatile Audio-Visual Learning for Emotion Recognition}

\author{
  Lucas Goncalves, Seong-Gyun Leem, Wei-Cheng
Lin, Berrak Sisman, Carlos Busso \\
  Multimodal Signal Processing (MSP) Lab \\
  Erik Jonsson School of Engineering and Computer Science \\
  The University of Texas at Dallas\\
  \texttt{\{goncalves, seong-gyun.leem, wei-cheng.lin, berrak.sisman, busso\}@utdallas.edu} \\
}

\begin{document}
\maketitle

\begin{abstract}

Most current audio-visual emotion recognition models lack the flexibility needed for deployment in practical applications. We envision a multimodal system that works even when only one modality is available and can be implemented interchangeably for either predicting emotional attributes or recognizing categorical emotions. Achieving such flexibility in a multimodal emotion recognition system is difficult due to the inherent challenges in accurately interpreting and integrating varied data sources. It is also a challenge to robustly handle missing or partial information while allowing direct switch between regression or classification tasks. This study proposes a \emph{versatile audio-visual learning} (VAVL) framework for handling unimodal and multimodal systems for emotion regression or emotion classification tasks. We implement an audio-visual framework that can be trained even when audio and visual paired data is not available for part of the training set (i.e., audio only or only video is present). We achieve this effective representation learning with audio-visual shared layers, residual connections over shared layers, and a unimodal reconstruction task. Our experimental results reveal that our architecture significantly outperforms strong baselines on the CREMA-D, MSP-IMPROV, and CMU-MOSEI corpora. Notably, VAVL attains a new state-of-the-art performance in the emotional attribute prediction task on the MSP-IMPROV corpus. 
\end{abstract}


\section{Introduction}
\label{sec:introduction}


%
%
%
%

Effective human interactions often include the expression and perceptions of emotional cues conveyed across multiple modalities to accurately comprehend and convey a message. During human interactions, there are two modalities that stand out as particularly significant: speech and facial expressions. These modalities play a crucial role in facilitating communication, making it essential for emotion recognition systems to incorporate speech-based cues \cite{Schuller_2018} and facial expression-based cues \cite{Tian_2001,Mariooryad_2016}. These modalities are intrinsically connected \cite{Busso_2006}, proving complementary information \cite{Busso_2004}. Therefore, emotion recognition systems can be more accurate if they incorporate audio-visual solutions \cite{DMello_2015, Khare_2021, Bouali_2022, Tran_2022}, mirroring the way humans interact in natural, real-world settings.

It is important to note that while humans primarily rely on these two modalities to recognize emotional states, there are scenarios in which only one modality may be utilized. In some cases, visual cues might be unavailable or occluded, leaving individuals to rely solely on acoustic information. Conversely, there might be situations where acoustic cues are insufficient or absent, requiring the exclusive use of visual cues for effective communication. As a result, it is vital for \emph{human computer interaction} (HCI) systems to not only be capable of simultaneously handling both modalities, but also to adapt to unimodal situations. The straightforward approach is to have separate unimodal and multimodal solutions. However, it is more computationally effective to have versatile systems that can be adapted according to the available information. This flexibility ensures that HCI systems can accommodate a wide range of communication contexts and maintain their effectiveness in various real-world conditions.

\emph{Artificial intelligent} (AI) methods have been explored for audio-visual learning to achieve high performance in either multimodal or unimodal scenarios \cite{Zhu_2021_2,Parthasarathy_2020_2, Goncalves_2022_3}. Most conventional models rely on building separate modality-specific branches \cite{Ngiam_2011, arandjelovic_2017, Antoniadis_2021} or employing ensemble-like techniques \cite{Hao_2020}, which can lead to convoluted and complex architectures. With the introduction of transformer frameworks \cite{Vaswani_2017}, many solutions have been developed to avoid training with modality-specific strategies. These models even implement formulations that unify the cross-modality relationships \cite{Akbari_2021,gong_2022} into a single, comprehensive model.

Despite significant recent advancements in the construction of simpler multimodal models, there are still open research questions to be considered when constructing a unified multimodal model. One of these challenges is that current methods tend to solely focus on one type of \emph{machine learning} (ML) task \cite{Baltrusaitis_2019}. For example, they consider either classification or regression problems, without considering the potential need for versatility across different problem types. In emotion recognition, this problem is especially important, since emotions can be alternatively described with emotional attributes (e.g., arousal, valence, dominance) and categorical emotions (e.g., anger, happiness, sadness) \cite{ElAyadi_2011}. 

Depending on the setting, therefore, it is important to have a model that can be utilized for either regression (emotional attributes) or classification (categorical emotions) settings without making major changes in the architecture.
Moreover, current methods often disregard the importance of maintaining distinct representations for each modality to capture modality-specific \cite{Zadeh_2018_2} features and characteristics within shared layers, which could affect performance in unimodal settings\cite{Ngiam_2011}. By failing to account for the unique characteristics of each modality, the models may not optimally leverage the strengths of each input type, ultimately limiting their effectiveness. Moreover, the complexity of many existing models can be a drawback, as it may lead to increased computational demands and reduced interpretability \cite{agrawal_2016}, making it more challenging for researchers and practitioners to analyze and adapt the models for various applications.  

Our main contribution in this study is the proposal of a \emph{versatile audio-visual learning} (VAVL) model, which unifies multimodal and unimodal learning in a single framework that can be used for emotion regression and classification tasks. Our approach utilizes branches that separately process each modality. In addition, we introduce shared layers, residual connections over shared layers, and a unimodal reconstruction task. The shared layers encourage learning representations that reflect the connections between the two modalities. The addition of the auxiliary reconstruction task and unimodal residual connections over the shared layers helps the model learn representations that reflect the heterogeneity of the modalities. Collectively, these components are added to our framework to help with the multimodal feature representation, which is a core challenge in multimodal learning \cite{Liang_2022}. We implement this audio-visual framework so it can be trained even when audio and visual paired data is not available for part of the data. The proposed approach is attractive as it enables the training of multimodal systems using incomplete information from multimodal databases (e.g., audio-only or visual-only information is available for some data points), as well as unimodal databases. The proposed framework also has the versatility to be used for either regression or classification tasks without changing the architecture or training strategy.

We quantitatively evaluate and compare the performance of our proposed model against strong baselines. The results demonstrate that our architecture achieves significantly better results on the CREMA-D \cite{Cao_2014_2}, the CMU-MOSEI \cite{Zadeh_2018_2}, and the MSP-IMPROV \cite{Busso_2017} corpora compared to the baselines. For example, VAVL achieves a new state-of-the-art result for the emotional attribute prediction task on the MSP-IMPROV corpus. Also, our architecture is able to sustain strong performance in audio-only and video-only settings, compared to strong unimodal baselines. Additionally, we conduct ablation studies into our model to understand the effects of each component on our model's performance. These results show the benefits of our proposed framework for audio-visual emotion recognition.

\section{Background}
\label{sec:related}

Emotion recognition has been widely explored using speech \cite{Schuller_2018, Parthasarathy_2017, Lotfian_2018}, facial expressions \cite{savchenko_2022}, and multimodal solutions \cite{Lin_2023_3,Li_2023, Yu_2023, Goncalves_2023}. However, most of these models have focused on solving either emotion classification tasks \cite{Goncalves_2023} or emotional attribute prediction tasks \cite{Parthasarathy_2017}. Furthermore, these studies are designed for either unimodal or multimodal solutions. Although some recent studies \cite{Antoniadis_2021} have proposed approaches for audio-visual emotion recognition that can handle both tasks, they still rely on a single modality. There is a need for solutions that can handle multiple modality settings. Recent advancements in deep learning have resulted in the development of unified frameworks that aim to move away from unimodal implementations. This section focuses on relevant studies that address similar problems investigated in this paper. 

\subsection{Formulations for Emotion Recognition}
\label{ssec:formulations}

Although an individual's cultural and ethnic background may impact her/his manner of expression, studies have indicated that certain basic emotions are very similar regardless of cultural differences \cite{Ekman_1971}. These basic emotions are described as: happiness, sadness, surprise, fear, disgust, and anger. These basic emotional states are often the most widely explored task in emotion recognition, formulating the problem as a six-class classification problem \cite{Goncalves_2022_2, ghaleb_2020}. In fact, many of the affective corpora include all or a subset of these classes, including the CREMA-D \cite{Cao_2014_2}, AffectNet \cite{Mollahosseini_2019}, and AFEW \cite{Dhall_2012} databases.

An alternative way to describe emotions is to leverage the continuous space of emotional attributes or sentiment analysis. The emotional attributes approach identifies dimensions to describe expressive behaviors. The most common attributes are arousal (calm versus active), valence (negative versus positive), and dominance (weak versus strong) \cite{Bradley_1994}. Several databases have explored the emotional attribute or sentiment analysis annotation of audio-visual stimuli, such as the IEMOCAP \cite{Busso_2008_5}, MSP-IMPROV \cite{Busso_2017}, and CMU-MOSEI \cite{Zadeh_2018_2} corpora. Studies have proposed algorithms to predict emotional attributes using audio-visual stimuli \cite{Schoneveld_2021, atmaja_2020, Hsu_2023}, facial expression \cite{wasi_2023}, and speech \cite{Schuller_2018, Wagner_2022,Upadhyay_2023_2}.

\subsection{Unified Multimodal Framework}
\label{ssec:uni_model}

Since the introduction of the transformer framework \cite{Vaswani_2017}, many studies have explored multimodal frameworks that can still work even when only one modality is available \cite{jaegle_2021}. Some proposed architectures employ training schemes that involve solely utilizing separate branches for different modalities, as seen in Baevski \etal \cite{baevski_2022}. These approaches construct branches containing unimodal information specific to each modality. In contrast, other studies explore the use of shared layers that encapsulate cross-modal information from multiple modalities \cite{dai_2022, shvetsova_2022}. The aforementioned studies incorporate aspects that laid the foundation for modeling a more unified framework, but they still have some limitations. For example, common solutions require independent modality training, assume that all the modalities are available during inference, or fail to fully address the heterogeneity present in a multimodal setting. Recent studies have proposed audio-visual frameworks that can handle both unimodal and multimodal settings \cite{Goncalves_2022}, but they still rely on having additional unimodal network branches. Gong \etal \cite{gong_2022} proposed a unified audio-visual framework for classification, which incorporates some effective capabilities, such as independently processing audio and video, and including shared audio-visual layers into the model. However, the model focused only on a classification task.

\subsection{Versatile Task Modeling}
\label{ssec:Versatile}

For audio-visual unified models \cite{gong_2022} and audio-visual models that have multiple modality branches and can handle unimodal scenarios \cite{Goncalves_2022_3}, the final multimodal prediction is often obtained by averaging the outputs from different branches, receiving the contributions from all the modalities. In emotion classification tasks, the model output is a probability distribution over a set of discrete categories. Averaging predictions from multiple unimodal branches can help capture different aspects of the data, leading to improved classification accuracy. However, this approach may not be effective for emotion attribute regression tasks, as regression involves predicting a continuous numerical value rather than a discrete categorical emotion label. Therefore, the average of the predictions from different unimodal branches may not be appropriate. This is particularly true when considering the performance gap that often exists between speech and visual models for predicting arousal, valence, and dominance, as facial-only features are generally less effective than acoustic features \cite{Parthasarathy_2020_2}. Simply averaging these two results could lead to suboptimal outcomes. A potentially appealing alternative is to use a weighted combination of the predictions \cite{Guo_2018, Khan_2018}. However, this approach also has its limitations, as the prediction discrepancy may not be consistent across all data points. To address this issue, our study proposes training an audio-visual prediction layer, in addition to visual and acoustic prediction layers, used when only data from one modality is available. The audio-visual prediction layer is optimized only when both modalities are available. This layer can automatically learn appropriate weights to combine the modality representations and fuse them to generate a single audio-visual prediction. This approach works well for both emotion classification and regression settings.

\subsection{Relation to Prior Work}
\label{ssec:relation}

While the implementation of our approach is with transformers, which have been used in the past in multimodal processing, our proposed VAVL model represents a significant contribution in comparison to previous works. The proposed architecture with shared layers, residual connections, reconstruction auxiliary tasks and separate audiovisual prediction layers is novel, where each of its components is carefully motivated to address a fundamental challenge in multimodal machine learning. We employ conformer layers \cite{gulati_2020} as our encoders, which are transformer-based encoders augmented with convolutions. The closest study to our framework is the work of Gong \etal \cite{gong_2022}, which proposed a unified audio-visual model for classification. Their framework involved the independent processing of audio and video features, with shared transformer and classification layers for both modalities. Prior to the shared transformer layers, they have modality-specific feature extractors and optional modality-specific transformers for each modality. In contrast, our framework employs conformer layers instead of vanilla transformers. Moreover, we introduce three important components to our framework. First, an audio-visual prediction layer is utilized solely when audio-visual modalities are available during inference, enhancing prediction accuracy and model versatility for both regression or classification tasks. Second, a unimodal reconstruction task is implemented during training to ensure that the shared layers capture the heterogeneity of the modalities. Third, residual connections are incorporated over the shared layers to ensure that the unimodal representations are not forgotten in the shared layers. This strategy maintains high performance even in unimodal settings. These novel additions represent significant contributions to the field of audio-visual emotion recognition. Additionally, our proposed model comprises approximately 86 million parameters. In comparison, TLSTM has 485,000 parameters, SFAV has 642,000 parameters, MulT has 749,000 parameters, and Auxformer has 1,226,000 parameters. While our model is more complex than these earlier models, it offers a significant reduction in complexity compared to the UAVM framework, which has 117 million parameters. This comparison indicates a reduction in complexity of approximately 26\% relative to UAVM, highlighting the efficiency of our model in handling sophisticated multi-modal tasks.

The proposed approach is significantly difference from our previous work on audio-visual emotion recognition \cite{Goncalves_2022_3}. Goncalves and Busso \cite{Goncalves_2022_3} proposed a method that incorporates two main components: the use of auxiliary unimodal networks and the use of modality dropout during training. The auxiliary networks ensure unimodal representations are separately embedded and not lost during the training of shared spaces within the cross-modal layers. The use of modality dropout enhances the ability of the model to retain performance when parts of the input data are missing. This model uses audiovisual paired data for training and performs averaging of prediction heads to obtain final predictions. In contrast, our approach utilizes distinct branches for processing each modality and introduces shared layers with residual connections, complemented by an unimodal reconstruction task. These shared layers are pivotal in facilitating the learning of representations that capture the interplay between audio and visual modalities. The shared layers take either visual or acoustic information, but not both, providing rich information to leverage cross-modality information. Furthermore, the incorporation of an auxiliary reconstruction task alongside unimodal residual connections enhances the model's ability to understand the diverse characteristics inherent in each modality. A key innovation of our approach lies in its ability to effectively train with incomplete multimodal data, accommodating scenarios where only audio or visual information is available for certain data points. This flexibility addresses a significant challenge in multimodal learning and broadens the applicability of our model, making it a substantial contribution to the field. The approach allows us to use partial audio or visual information from multimodal and unimodal databases.

In summary, in relation to previous works, the primary novelty of our framework lies within the unique organization and training methodology of its layers. Our approach strategically structures its layers in a manner that optimizes their functionality and efficacy for our specific application for audio-visual and unimodal (visual or acoustic) emotion recognition. 

\begin{figure}[t]
    \centering
    \includegraphics[width=85mm,height=85mm,scale=0.5]{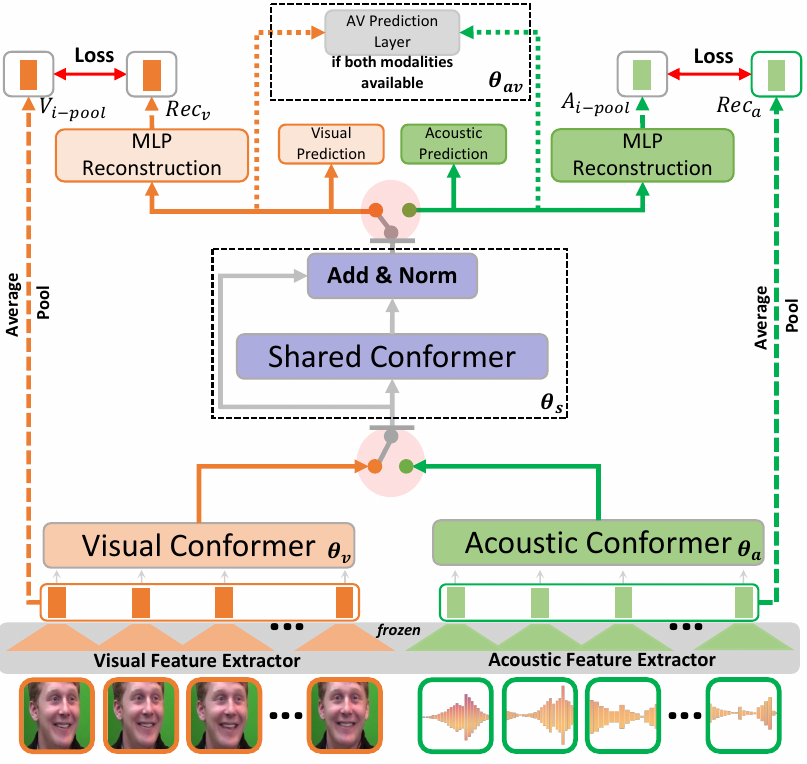}
    \caption{Overview of our proposed \emph{versatile audio-visual learning} (VAVL) framework. The orange branches represent the visual information and are parameterized by the weights $\theta_v$. The green branches represent the acoustic information and are parameterized by the weights $\theta_a$. The purple modules are the shared layers where both modalities flow through, which are parameterized by the weights $\theta_s$. Finally, the gray module is the audio-visual prediction layer, which is parameterized by the weights $\theta_{av}$.}
    \label{fig:model}
\end{figure}

\section{Proposed Approach}
\label{sec:ProposedApproach}

Figure \ref{fig:model} illustrates the architecture of the proposed approach, which consists of four major components. Each of these components is represented in the figure by the set of weights $\theta_v$, $\theta_a$, $\theta_s$, and $\theta_{av}$. The orange branches represent parts of the model where only visual information flows. This region is parameterized by the weights $\theta_v$, and includes the visual conformer encoder layers that encode the visual input representations, the visual prediction layer for visual-only predictions, and the MLP reconstruction layers for the averaged visual input features. The green branches represent parts of the model where only acoustic information flows. This region is parameterized by the weights $\theta_a$, and consists of the acoustic conformer encoder layers that encode the acoustic input representations, the acoustic prediction layer for acoustic-only predictions, and the MLP reconstruction layers for the averaged acoustic input features. The shared layers, depicted in purple in Figure \ref{fig:model}, are parameterized by the weights $\theta_s$. This block processes both acoustic and visual inputs to learn intermediate audio-visual representations from both modalities. It also contains residual connections from the unimodal branches to the shared layers to preserve information from the unimodal representations. Lastly, the audio-visual prediction layer is parameterized by the weights $\theta_{av}$, which focuses on processing audio-visual representations from the shared layers when the model receives paired audio-visual data for training or inference. The following sections describe these components in detail.

\subsection{Acoustic and Visual Layers}
\label{ssec:acousticvisual}

As shown in Figure \ref{fig:model}, the acoustic and visual layers mirror each other. Both layers share the same basic structure and training mechanism. The main components of these layers are conformer encoders \cite{gulati_2020}, which process all the sequential video or acoustic frames in parallel, depending on the modality available during either training or inference. To obtain inputs for our model, we utilize pre-trained feature extractors that produce a 1,408D feature vector for each visual frame and a 1,024D feature vector for each acoustic frame. Section \ref{sec:features} describes these feature extractor modules in detail. We apply a 1D temporal convolutional layer at the frame level before entering the feature vectors to their corresponding conformer encoder layers. This step ensures that each element in the input sequences is aware of its neighboring elements. This approach also allows us to project both feature vectors into a 50D feature representation, matching their dimensions. The dimensionality of the projection was determined based on preliminary experiments. These experiments indicated that a 50D representation strikes an effective balance: it retains essential information from the original feature vectors while keeping the model's complexity manageable.

As seen in the top region of Figure \ref{fig:model}, there are two additional components separately implemented for each modality (acoustic and visual) that are implemented to (1) predict the emotional attributes, and (2) reconstruct the unimodal feature representations. For the prediction of the emotional attributes, the acoustic and visual layers contain a set of prediction layers, referred to as visual prediction and acoustic prediction in Figure \ref{fig:model}. They are responsible for generating unimodal predictions, which are utilized when our model operates in a unimodal setting. For the reconstruction of the unimodal feature representation, the model has a \emph{multilayer perceptron} (MLP) head that serves as an auxiliary reconstruction task. The reconstruction task is used to have the model reconstruct the average-pooled input representations for the modality being used during training at that moment. The input of the reconstruction is the average-pooled representations obtained at the output of the shared layers (Sec. \ref{ssec:SharedLayers}). Equation \ref{eq_totalloss} shows the total loss function of the model,

 \begin{equation}\label{eq_totalloss}
    \mathcal{L}= \mathcal{L}_{pred}(y, pred) + \alpha\mathcal{L}_{MSE}(x_{pool}, Rec_M) 
 \end{equation}

\noindent
where $\mathcal{L}_{pred}$ is the emotion task-specific loss (i.e., cross-entropy -- Eq. \ref{eq_ce}, or \emph{concordance correlation coefficient} (CCC) -- Eq. \ref{eq_ccc}), $y$ is the input label, $pred$ is the emotional prediction of the model, $\alpha$ is the scaling weight for the reconstruction loss, $\mathcal{L}_{MSE}$ is \emph{mean squared error} (MSE) loss between the average-pooled input value $x_{pool}$ and the reconstructed input $Rec_M$ for modality $M$. The terms $x_{pool}$ and $Rec_M$ are specific to each modality; $x_{pool}$ represents the pooled features of the input (audio or video), while $Rec_M$ represents the reconstructed features for the corresponding modality. This differentiation ensures that the reconstruction loss appropriately guides the model to retain and reconstruct modality-specific information.

The reconstruction task is included in our model to promote the learning of more general features that can be applied to both modalities while preserving separate information for each modality, as the reconstruction from the shared layers requires the model to retain information from both modalities.

\subsection{Shared Layers}
\label{ssec:SharedLayers}

The shared layers, depicted in purple in Figure \ref{fig:model}, mainly comprise a conformer encoder, following a similar structure to the acoustic and visual layers. These shared layers are modality-agnostic, meaning that during training and inference, the features from both modalities will pass through these layers whenever each modality is available. The purpose of this block is to ensure that the shared layers maintain information from both modalities incorporated in our model.

We also introduce a residual connection over the shared layers (gray arrow in model $\theta_s$ shown in Fig. \ref{fig:model}). This residual connection from the unimodal branches over the shared layers ensures that the model retains unimodal separable information. This mechanism complements the reconstruction tasks mentioned in Section \ref{ssec:acousticvisual} to increase the robustness of the system when only one modality is available. The approach allows the gradients to flow directly from the shared layers to the unimodal branches, preserving modality-specific information in our model.

\subsection{Audio-Visual Prediction Layer}
\label{ssec:AudiovisualModel}

When only one modality is available, the system will use the visual or acoustic prediction blocks. When both modalities are available for training or inference, we use the audio-visual prediction layer, highlighted in gray in Figure \ref{fig:model}. This layer plays a crucial role in making our approach versatile. Unlike other methods that rely on averaging predictions from different branches in their models, the audio-visual layers effectively utilize the contributions of acoustic and visual inputs in the audio-visual space. The audio-visual prediction layer consists of two fully connected layers that feed into a final audio-visual prediction head. We added this simple structure to ensure that representations from both modalities, obtained from the average-pooled output of each modality from the shared layers, are properly combined to obtain a final audio-visual prediction. During training, when audio-visual data is available, all other layers of the model are frozen after updating with acoustic and visual data. Only this layer is separately optimized to learn the required weights for audio-visual predictions. This layer ensures proper combination of audio-visual representations for robust predictions in both classification or regression settings.

\begin{algorithm} [t]
\caption{- VAVL (Training and Inference)}
\begin{algorithmic}[1]
\REQUIRE $\mathcal{D} = \left\{A, V, \textit{Label}\right\}$, $\mathcal{M} = \left\{\theta_a, \theta_v, \theta_s, \theta_{av}\right\}$
\vspace{0.1cm}
\hrule
\vspace{0.1cm}
\hspace*{-2em} $\textbf{Training} (\mathcal{D}, \mathcal{M}, \alpha$)
\vspace{0.1cm}
\hrule
\vspace{0.1cm}
\WHILE{$i < \textit{max train iterations}$}
    \STATE sample a batch of $\left\{A_i, V_i, Label_i\right\}$
    \vspace{0.1cm}
    \IF{$A_i \neq \text{None}$}
        \STATE{$Pred_a, A_{i-pool}, Rec_a = \mathcal{M}_{\left\{\ \theta_a, \theta_s\right\}}(A_i)$}
        \STATE{$\mathcal{L}= \mathcal{L}_{pred}(Label_i, Pred_a) + \alpha\mathcal{L}_{MSE}(A_{i-pool}, Rec_a) $} 
        \STATE backpropagate and update $\left\{\theta_a, \theta_s\right\}$
    \ENDIF
    \vspace{0.1cm}
    \IF{$V_i \neq \text{None}$}
        \STATE{$Pred_v, V_{i-pool}, Rec_v = \mathcal{M}_{\left\{\ \theta_v, \theta_s\right\}}(V_i)$}
        \STATE{$\mathcal{L}= \mathcal{L}_{pred}(Label_i, Pred_v) + \alpha\mathcal{L}_{MSE}(V_{i-pool}, Rec_v) $} 
        \STATE backpropagate and update $\left\{\theta_v, \theta_s\right\}$
    \ENDIF
    \vspace{0.1cm}
    \IF{$A_i \neq \text{None}$ \AND $V_i \neq \text{None}$}
        \STATE $freeze$ $\left\{\theta_a, \theta_v, \theta_s\right\}$
        \STATE{$Pred_{av} = \mathcal{M}_{\left\{\theta_a, \theta_v, \theta_s, \theta_{av}\right\}}(A_i, V_i)$}
        \STATE{$\mathcal{L}= \mathcal{L}_{pred}(Label_i, Pred_{av}) $} 
        \STATE backpropagate and update $\left\{\theta_{av}\right\}$
    \ENDIF
\ENDWHILE
\vspace{0.1cm}
\STATE \textbf{return} $\mathcal{M}$
\vspace{0.1cm}
\hrule
\vspace{0.1cm}
\hspace*{-2em} \textbf{Inference} ($\left\{A,V\right\}, \mathcal{M}$)
\vspace{0.1cm}
\hrule
\vspace{0.1cm}
\IF{$A \neq \text{None}$ \AND $V \neq \text{None}$}
    \STATE{$Pred = \mathcal{M}_{\left\{\theta_a, \theta_v, \theta_s, \theta_{av}\right\}}(A, V)$}
\vspace{0.1cm}
\ELSIF{ one modality is not available}
\STATE{$Pred = \mathcal{M}_{\left\{\ \theta_a, \theta_s\right\}}(A)$} when $V == None$
\STATE{$Pred = \mathcal{M}_{\left\{\ \theta_v, \theta_s\right\}}(V)$} when $A == None$
\ENDIF
\STATE \textbf{return} $Pred$
\end{algorithmic}
\label{alg:VAVL}
\end{algorithm}

\subsection{Versatile Model Training}
\label{ssec:training}

We train our proposed VAVL model using Algorithm \ref{alg:VAVL}. The training process can involve either a single modality or both modalities at any given iteration. If both modalities are available, the model first backpropagates the errors using the acoustic predictions to optimize the acoustic and shared weights ($\theta_a$, $\theta_s$). Then, it backpropagates the errors using the visual predictions to optimize the visual and shared weights ($\theta_v$, $\theta_s$). Next, it freezes all the parameters of the models other than the audio-visual prediction block (i.e., $\theta_v$, $\theta_a$, and $\theta_s$ are frozen). Then, it backpropagates the errors using the audio-visual predictions to optimize the audio-visual prediction weights ($\theta_{av}$). This method facilitates the use of unpaired and paired audio-visual data for training. If only one modality is available, we only update either the visual and shared weights ($\theta_v$, $\theta_s$) or acoustic and shared weights ($\theta_a$, $\theta_s$). Essentially, most of the blocks in the proposed architecture are trained with either acoustic features or visual features. The only block that requires paired data is the audio-visual prediction layer. The shared conformer layer processes each modality separately when full audiovisual information is available, so the dimension of the input to this block has always consistent dimension. The outputs are then combined in the audio-visual prediction layer.

At inference time, the model utilizes the available information to make predictions, and in cases where both modalities are present, the model outputs the predictions from the audio-visual prediction layer. When only one modality is available during inference, the model outputs the prediction from the corresponding visual or acoustic prediction blocks.

\subsection{Complexity Analysis}
\label{ssec:complexity}

An analysis of the complexity of the VAVL method reveals a framework with 86M trainable parameters, underscoring its capacity for handling complex multi-modal tasks. Central to the framework, we have the acoustic and visual conformers, each equipped with 34M parameters, adeptly processing audio and visual inputs. The shared conformer, integral to the model's functionality, contains 15M parameters, facilitating seamless integration of different modalities. A crucial aspect of our model is the 1D temporal convolutional layers, containing 650k parameters. These blocks are important in projecting both visual and acoustic features into a representation with the same dimension. Additionally, the model features three specialized prediction heads for acoustic, audio-visual, and visual processing, each with 390K parameters, further enhancing its processing capabilities. Lastly, the reconstruction MLP layers contain a combined total of 1.4 million parameters. To evaluate the practical efficiency, we measured the inference latency and \emph{real-time factor} (RTF) on a 3090 RTX GPU. The results are as follows: audio-only mode had an average latency of 0.003499 seconds (RTF: 0.000005), video-only mode had an average latency of 0.001759 seconds (RTF: 0.000003), and audiovisual mode had an average latency of 0.003770 seconds (RTF: 0.000006).

\section{Experimental Settings}
\label{sec:experiments}

\subsection{Emotional Corpora}
\label{ssec:corpora}

In this study, we use the CREMA-D \cite{Cao_2014_2}, the MSP-IMPROV \cite{Busso_2017}, and the CMU-MOSEI \cite{Zadeh_2018_2} corpora. The CREMA-D corpus is an audio-visual corpus with high-quality recordings from 91 ethically and racially diverse actors (48 male, 43 female). Actors were asked to convey specific emotions while reciting sentences. Videos were recorded against a green screen background, with two directors overseeing the data collection. One director worked with 51 actors, while the other worked with 40 actors. Emotional labels were assigned by at least seven annotators. In total, 7,442 clips were collected and rated by 2,443 raters. We use the perceived emotions from the audio-visual modality in the CREMA-D corpus for our classification task. We consider six emotional classes: anger, disgust, fear, happiness, sadness, and neutral state.

The MSP-IMPROV corpus \cite{Busso_2017} is the second audio-visual database used in this study. The corpus was collected to study emotion perception. The corpus required sentences with identical lexical content but conveying different emotions. Instead of actors reading sentences, a sophisticated protocol elicited spontaneous target sentence renditions. The corpus includes 20 target sentences with four emotional states, resulting in 80 scenarios and 652 speaking turns. It also contains the interactions that prompted the target sentence (4,381 spontaneous speaking turns), natural interactions during breaks between the recording of dyadic scenarios (2,785 natural speaking turns), and read recordings expressing the target emotional classes (620 read speaking turns). In total, the MSP-IMPROV corpus contains 7,818 non-read speaking turns and 620 read sentences. The corpus was annotated using a crowdsourcing protocol, monitoring the quality of the workers in real-time. Each sentence was annotated for arousal (calm versus active), valence (negative versus positive), and dominance (weak versus strong) by five or more raters using a five-point Likert scale. We employ these three emotional attributes for our regression task.

The CMU-MOSEI \cite{Zadeh_2018_2} comprises review video clips of movies sourced from YouTube. Each clip is annotated by human experts with a sentiment score ranging from -3 to 3. We retrieved the data from the author's SDK and obtained a total of 22,859 files of which, based on their standard splits, 16,326 are used for training, 1,871 are used for development, and 4,659 are used for testing. This database consists of in-the-wild audio-visual recordings.

\subsection{Acoustic and Visual Features}
\label{sec:features}

For the CREMA-D and MSP-IMPROV corpora, we have access to the raw videos and audio recordings, so we can extract audio and visual features. Our acoustic feature extractor is based on the ``wav2vec2-large-robust" architecture \cite{Hsu_2021_2}, which has shown superior emotion recognition performance compared to other variants of the Wav2vec2.0 model \cite{Baevski_2020}, as demonstrated in the study by Wagner \etal \cite{Wagner_2022}. The downstream head of our model consists of two fully connected layers with 1,024 nodes, layer normalization, and \emph{rectified linear unit} (ReLU) activation function, followed by a linear output layer with three nodes for predicting emotional attribute scores (arousal, dominance, and valence). We import the pre-trained ``wav2vec2-large-robust" model from the Hugging Face library \cite{Wolf_2019}. We use this wav2vec2 model specifically pre-trained for emotion recognition tasks before its integration to ensure the representations used are optimal for emotion recognition. We aggregate the output of the transformer encoder using average pooling and feed them to the downstream head. To regularize the model and prevent overfitting, we utilize dropout with a rate of $p=0.5$ applied to all hidden layers. To fine-tune the model, we use the training set of the MSP-Podcast corpus (release of v1.10) \cite{Lotfian_2019_3}. The ADAM optimizer \cite{Kingma_2014_2} is employed with a learning rate set to 0.0001. We update the model with mini-batches of 32 utterances for 10 epochs. With the fine-tuned ``wav2vec2-large-robust” model, we extract acoustic representations with a 25ms window size and a 20 ms stride from the given audio, which are then used as the acoustic features. This strategy creates 50 frames per second. 

To obtain visual features, we used the \emph{multi-task cascaded convolutional neural network} (MTCNN) face detection algorithm \cite{Zhang_2016_8} to extract faces from each image frame in the corpora using bounding boxes. Following the extraction of bounding boxes, we resize the images to a predetermined dimension of $224 \times 224 \times 3$. After face extraction, we utilize the pre-trained EfficientNet-B2 model \cite{savchenko_2022} to extract emotional feature representations. At the time of our study, this model was among the top performers on the AffectNet corpus \cite{Mollahosseini_2019}. Similarly to the acoustic feature extractor, the EfficientNet-B2 model is pre-trained for emotion recognition tasks before its integration. This approach helps ensure the representations obtained from videos are optimal for emotion recognition. The representation obtained from the EfficientNet-B2 model is retrieved from the last fully connected layer before classification, with an array dimension of 1,408. This representation is then concatenated row-wise with all other frames within each clip from the datasets, serving as the input for the visual branch of our framework.

The CMU-MOSEI dataset offers pre-extracted features instead of raw data. Specifically, for the visual modality, it provides 35 facial action units, while the audio data includes features such as \emph{Mel-frequency cepstral coefficients} (MFCCs), pitch tracking, glottal source, and peak slope parameters, totaling 74 features. Consequently, we could not use the same set of features used to conduct experiments on the CREMA-D and MSP-IMPROV corpora (i.e., wav2vec2 and EfficientNet-B2 based features). Instead, we use the features provided with the release, which also facilitates the comparison with other methods using this corpus.

\subsection{Implementation Details}
\label{ssec:implementation}

We implemented the conformer blocks with an encoder hidden layer set to 512D, with 8 attention heads. We set a dropout rate to $p=0.1$. The number of layers in the acoustic, visual, and shared conformer layers were set to three, three, and two, respectively. The acoustic/visual prediction layers and the reconstruction layers are implemented using a fully-connected structure with three layers. The first two layers are implemented with 512 nodes and 256 nodes, respectively. For the acoustic and visual prediction modules, the third layer is the output layer, where the size depends on the emotion recognition task (i.e., classification or regression). For the reconstruction task, the third layer is the target representation to be reconstructed, so it has 1,024 nodes for the acoustic features and 1,408 nodes for the visual features. The fully connected layers are implemented with dropout, with the rate set to $p=0.2$. Similarly, the audio-visual prediction layers have a mostly identical fully-connected hidden layer structure to the unimodal prediction layers. The only difference is that the input layers are now 1,024D, as they need to take both unimodal representations from the shared layers as inputs. We trained the models for 20 epochs using the ADAM optimizer, with the ReLU as the activation function. The learning rate is set to 5e-5. We used a batch size of 32 for the CREMA-D and CMU-MOSEI dataset experiments, and a batch size of 16 for the MSP-IMPROV dataset experiments, since the MSP-IMPROV has longer sentences. The model was implemented in PyTorch and trained using an Nvidia Tesla V100.

The recordings in each database are divided into train, development, and test sets, with approximately 70\%, 15\%, and 15\% of the data in each set, respectively. The splits were carried out in a speaker-independent manner, ensuring that no speaker appeared in more than one set. We trained each model five times, with different splits each time. All models were trained using the train set, and the best-performing model on the development set was selected and used to make predictions on the test set.

\subsection{Cost Functions and Evaluation Metrics}
\label{ssec:metrics}

Three of the most prominent tasks in affective computing are classification tasks for categorical emotions, regression models for emotional attributes, and sentiment analysis which can be approached as either a classification or a regression task, depending on the specifics of the problem and the nature of the sentiment scores. Our model is designed to be utilized and evaluated in both formulations (i.e., classification or regression tasks).

For emotion classification, our training objective is based on the multiclass cross-entropy loss, as seen in equation \ref{eq_ce},

 \begin{equation}\label{eq_ce}
    \mathcal{L}_{CE} = -\sum_{c=1}^My_{o,c}\log(p_{o,c})
 \end{equation}

\noindent
where $M$ is number of classes, $log$ is the natural log, $c$ is the correct label for observation $o$, and $p$ is the predicted probability that observation $o$ belong to class $c$. We predict the emotional state/value for each input sequence in our test set and report the `micro' and `macro' F1-scores. The `micro' F1-score is computed by considering the total number of true positives, false negatives, and false positives, making it sensitive to class imbalance. The `macro' F1-score calculates the F1-score for each class separately and aggregated the scores with equal weight, so the performance on the minority class is as important as the performance in the majority class. 

For emotional attributes regression models, we use the \emph{concordance correlation coefficient} (CCC), which measures the agreement between the true and predicted emotional attribute scores. Equation \ref{eq_ccc} illustrates the CCC measurements,

\begin{equation}\label{eq_ccc}
    \mathcal{L}_{CCC} = \frac{2\rho\sigma_x\sigma_y}{\sigma_x^2+\sigma_y^2+(\mu_x-\mu_y)^2}
 \end{equation}

\noindent 
where $\mu_x$ and $\mu_y$ are the means of the true and predicted scores, $\sigma_x$ and $\sigma_y$ are the standard deviation of the true and predicted scores, and $\rho$ is their Pearson's correlation coefficient. We train our model to maximize CCC so that the predicted scores have a high correlation with the true scores, while reducing their errors in the prediction. We use the CCC metric for model evaluation of the predictions of arousal, valence, and dominance.

We formulate the prediction of sentiment analysis as the prediction of a continuous number between -3 to 3. We employ the \emph{mean absolute error} (MAE) as the primary training objective. This \emph{L1 loss} function is defined as the average of the absolute differences between the predicted sentiment scores and the true sentiment scores. The L1 loss is formulated as shown in Equation \ref{eq_l1_loss}:

\begin{equation}\label{eq_l1_loss}
    \mathcal{L}_{L1} = \frac{1}{N} \sum_{i=1}^{N} |y_i - \hat{y}_i|
\end{equation}

\noindent
where $N$ is the number of samples, $y_i$ is the true sentiment score, and $\hat{y}_i$ is the predicted sentiment score for the $i$-th sample. The L1 loss is particularly suitable for sentiment analysis as it robustly handles outliers and ensures a linear error penalization. Our model aims to minimize this loss during training, thereby reducing the average magnitude of errors in sentiment score predictions. This loss function is instrumental in achieving high accuracy in predicting the nuanced sentiment scores across the specified range.

We conducted each experiment in this work five times with different partitions or seeds and reported the average metrics. Additionally, we performed statistical analyses to evaluate our model's performance against the baseline models. We used a two-tailed t-test, asserting a significance level at $p$-value$<$0.05.

\begin{table*}[ht]
\captionsetup{justification=centering}
\caption{Comparison between the VAVL model and the audio-visual and unimodal baselines. The table reports the average performance metrics across five trials. The symbol $\ast$ indicates that the VAVL model is significantly better than the other baselines on the CREMA-D and CMU-MOSEI datasets.}
\label{tab:performances_1}
\centering
\scriptsize 
\setlength{\tabcolsep}{2pt} 
\begin{tabularx}{\textwidth}{l *{6}{>{\centering\arraybackslash}X} | *{3}{>{\centering\arraybackslash}X}}
\toprule
& \multicolumn{6}{c|}{\textbf{CREMA-D}} & \multicolumn{3}{c}{\textbf{CMU-MOSEI}} \\
\cmidrule(lr){2-7} \cmidrule(lr){8-10}
 & \multicolumn{2}{c}{\textbf{Audio-Visual}} & \multicolumn{2}{c}{\textbf{Acoustic}} & \multicolumn{2}{c|}{\textbf{Visual}} & \textbf{Audio-Visual} & \textbf{Acoustic} & \textbf{Visual} \\
\cmidrule(lr){2-3} \cmidrule(lr){4-5} \cmidrule(lr){6-7} \cmidrule(lr){8-8} \cmidrule(lr){9-9} \cmidrule(lr){10-10}
\textbf{Model} & F1-Ma. & F1-Mi. & F1-Ma. & F1-Mi. & F1-Ma. & F1-Mi. & MAE & MAE & MAE \\
\midrule
VAVL & \textbf{0.779$\pm$.025} & \textbf{0.826$\ast\pm$.015} & 0.628$\pm$.013 & \textbf{0.701$\ast\pm$.015} & \textbf{0.738$\ast\pm$.033} & \textbf{0.787$\ast\pm$.020} & \textbf{0.792$\pm$.029} & 0.829$\pm$.034 & \textbf{0.795$\ast\pm$.028} \\
UAVM & 0.737$\pm$.018 & 0.804$\pm$.008 & 0.552$\pm$.034 & 0.607$\pm$.033 & 0.691$\pm$.017 & 0.748$\pm$.013 & 0.802$\pm$.029 & 0.862$\pm$.031 & 0.803$\pm$.026 \\
AuxFormer & 0.743$\pm$.019 & 0.791$\pm$.017 & 0.504$\pm$.084 & 0.578$\pm$.046 & 0.692$\pm$.033 & 0.739$\pm$.032 & 0.808$\pm$.030 & 0.830$\pm$.032 & 0.806$\pm$.028 \\
SFAV & 0.771$\pm$.020 & 0.810$\pm$.010 & \textbf{0.658$\pm$.015} & 0.699$\pm$.014 & 0.656$\pm$.019 & 0.709$\pm$.015 & 0.804$\pm$.024 & 0.821$\pm$.032 & 0.813$\pm$.032 \\
TLSTM & 0.667$\pm$.012 & 0.746$\pm$.012 & 0.503$\pm$.046 & 0.543$\pm$.042 & 0.574$\pm$.020 & 0.651$\pm$.025 & 0.811$\pm$.022 & 0.821$\pm$.032 & 0.815$\pm$.028 \\
MulT & 0.714$\pm$.015 & 0.762$\pm$.015 & ---- & ---- & ---- & ---- & 0.810$\pm$.028 & ---- & ---- \\
Uni. (A) & ---- & ---- & 0.625$\pm$.024 & 0.690$\pm$.015 & ---- & ---- & ---- & \textbf{0.821$\pm$.028} & ---- \\
Uni. (V) & ---- & ---- & ---- & ---- & 0.725$\pm$.039 & 0.783$\pm$.028 & ---- & ---- & 0.859$\pm$.032 \\
\bottomrule
\end{tabularx}
\end{table*}

\begin{table*}[ht]
\captionsetup{justification=centering}
\caption{Comparison between the VAVL model and the audio-visual and unimodal baselines. The table reports the average performance metrics across five trials. The symbol $\ast$ indicates that the VAVL model is significantly better than the other baselines on the MSP-IMPROV dataset.}
\label{tab:performances_2}
\centering
\scriptsize 
\setlength{\tabcolsep}{2pt} 
\begin{tabularx}{\textwidth}{l *{9}{>{\centering\arraybackslash}X}}
\toprule
& \multicolumn{9}{c}{\textbf{MSP-IMPROV}} \\
\cmidrule(lr){2-10}
& \multicolumn{3}{c}{\textbf{Audio-Visual}} & \multicolumn{3}{c}{\textbf{Acoustic}} & \multicolumn{3}{c}{\textbf{Visual}} \\
\cmidrule(lr){2-4} \cmidrule(lr){5-7} \cmidrule(lr){8-10}
\textbf{Model} & Aro. & Val. & Dom. & Aro. & Val. & Dom. & Aro. & Val. & Dom. \\
\midrule
VAVL & \textbf{0.856$\ast\pm$.110} & \textbf{0.876$\ast\pm$.095} & \textbf{0.814$\ast\pm$.151} & \textbf{0.853$\ast\pm$.104} & 0.858$\pm$.111 & 0.783$\pm$.155 & \textbf{0.422$\ast\pm$.133} & \textbf{0.631$\ast\pm$.032} & 0.375$\pm$.112 \\
UAVM & 0.471$\pm$.257 & 0.687$\pm$.329 & 0.544$\pm$.309 & 0.578$\pm$.316 & 0.705$\pm$.339 & 0.637$\pm$.384 & 0.274$\pm$.145 & 0.522$\pm$.260 & 0.296$\pm$.163 \\
AuxFormer & 0.672$\pm$.134 & 0.820$\pm$.071 & 0.652$\pm$.149 & 0.722$\pm$.218 & 0.789$\pm$.142 & 0.730$\pm$.162 & 0.363$\pm$.153 & 0.581$\pm$.071 & 0.293$\pm$.141 \\
SFAV & 0.786$\pm$.103 & 0.761$\pm$.119 & 0.721$\pm$.147 & 0.745$\pm$.081 & 0.699$\pm$.073 & 0.628$\pm$.140 & 0.345$\pm$.192 & 0.572$\pm$.175 & \textbf{0.473$\pm$.148} \\
TLSTM & 0.832$\pm$.095 & 0.843$\pm$.103 & 0.767$\pm$.155 & 0.835$\pm$.117 & 0.765$\pm$.188 & \textbf{0.841$\pm$.125} & 0.172$\pm$.140 & 0.574$\pm$.094 & 0.201$\pm$.066 \\
MulT & 0.775$\pm$.088 & 0.761$\pm$.056 & 0.778$\pm$.136 & ---- & ---- & ---- & ---- & ---- & ---- \\
Uni. (A) & ---- & ---- & ---- & 0.841$\pm$.095 & \textbf{0.878$\pm$.108} & 0.820$\pm$.158 & ---- & ---- & ---- \\
Uni. (V) & ---- & ---- & ---- & ---- & ---- & ---- & 0.383$\pm$.065 & 0.598$\pm$.093 & 0.321$\pm$.070 \\
\bottomrule
\end{tabularx}
\end{table*}
\subsection{Baselines}
\label{ssec:baselines}
To evaluate the performance of our proposed model, we conducted experiments with three strong audio-visual frameworks and one unimodal framework. Baseline models 1, 2, 3, and 4 were implemented utilizing the code available in their respective repositories. For Baseline 5, the implementation was carried out in accordance with the specifications outlined in its associated paper. Baseline 6 represents a unimodal variant of our model, specifically crafted for comparisons in a unimodal setting.

\subsubsection{Baseline 1: UAVM}
Gong \etal \cite{gong_2022} proposed a unified audio-visual framework for classification, which independently processes audio and video features. The framework includes a shared transformer component and a classification layer, whose weights are shared between the modalities.

\subsubsection{Baseline 2: AuxFormer}
Goncalves and Busso \cite{Goncalves_2022} proposed an audio-visual transformer-based framework that creates cross-modal representations through transformer layers, sharing representations from query inputs from one modality to the keys and values of another modality. Additionally, their architecture includes unimodal auxiliary networks and modality dropout during training to enhance the robustness to missing modalities, allowing the model to be used in both audio-visual and unimodal settings.

\subsubsection{Baseline 3: MulT}

Tsai \etal \cite{Tsai_2019} proposed a multimodal transformer architecture for human language time-series data. Their model uses a cross-modal transformer framework that generates pairs of bimodal representations, where the keys and values of one modality interact with the queries of a target modality. Vectors with similar target modalities are concatenated and passed to another transformer layer that generates representations used for prediction. The original model considered textual, visual, and acoustic features. We adapt the model from the original study by removing the dependencies on the textual branch, focusing only on visual and acoustic features. 

\subsubsection{Baseline 4: SFAV}

Chumachenko \etal \cite{Chumachenko_2022} present an architecture for audiovisual emotion recognition, particularly addressing the challenge of incomplete data from either modality during inference. Their model is designed to learn from both audio and visual data and incorporates robust fusion mechanisms that perform well even when one modality is absent. The authors explore fusion techniques, such as late transformer fusion and intermediate transformer fusion to more effectively integrate features from the audio and visual branches. In this paper, we refer to this baseline as \emph{SFAV} in agreement with their methodology approach which uses self-attention fusion for audio-visual emotion recognition.

\subsubsection{Baseline 5: TSLTM}

Huang \etal \cite{Huang_2020} propose a multimodal transformer architecture for continuous emotion recognition, leveraging the Transformer's ability to model long-term temporal dependencies with self-attention mechanisms. Their model combines audio and visual modalities through model-level fusion without an encoder-decoder structure. It employs multi-head attention to learn emotional temporal dynamics and fuses audio-visual modalities into a shared semantic space, outperforming traditional fusion methods. Additionally, the architecture integrates \emph{long short-term memory} (LSTM) networks to further improve performance. In our study, we refer to the model as \emph{TLSTM} consistent with their method of combining the transformer model and LSTM for emotion recognition.

\subsubsection{Baseline 6: Unimodal Acoustic and Visual Model}
The unimodal baseline model has a similar structure to the model proposed in this study. Like the acoustic and visual layers of our model, it uses conformer encoder layers \cite{gulati_2020} to process all sequential video or acoustic frames in parallel. The output of the conformer layers is then average-pooled and fed into a network that contains two fully connected layers to generate the prediction. We build the unimodal network with five conformer layers to match the structure of the full VAVL network.

\section{Experimental Results}
\label{ssec:experiments}

\subsection{Comparison with Baselines}

This section compares our proposed model with the audiovisual and unimodal baselines explored in this study. All models were trained using the details discussed in Section \ref{ssec:implementation}. Tables \ref{tab:performances_1} and \ref{tab:performances_2} presents the average results for all the models across the five trials on each corpus. The models' performances are evaluated based on three modalities: audio-visual, acoustic, and visual. For the MSP-IMPROV dataset, the performance is assessed using the CCC predictions for \emph{arousal} (Aro.), \emph{valence} (Val.), and \emph{dominance} (Dom.). For the CREMA-D dataset, we report the \emph{F1-Macro} (F1-Ma) and \emph{F1-Micro} (F1-Mi) scores. For the CMU-MOSEI dataset, we report \emph{Mean Absolute Error} (MAE). Upon examining the performance metrics presented in the tables, the VAVL model demonstrates notable superiority across various tasks when compared to the baseline models on CREMA-D, MSP-IMPROV, and CMU-MOSEI datasets. This superiority is quantified by the asterisks indicating statistical significance.

In the audio-visual modality on the CREMA-D dataset, the VAVL model achieves an F1-Macro score of 0.779$\pm$0.025, which is significantly higher than that of the strongest baseline, the TSLTM model, at 0.667$\pm$0.012. This pattern is consistent in the F1-Micro score, where VAVL scores 0.826$\pm$0.015, outperforming the second-best SFAV model's score of 0.810$\pm$0.010.

The VAVL model's performance in acoustic modality is also strong, with a MAE of 0.829$\pm$0.034 on the CMU-MOSEI dataset, which is lower than the best-performing baseline model (AuxFormer) with an MAE of 0.830$\pm$0.032. Lower MAE indicates better performance, and these numbers underscore the predictive accuracy of the VAVL model in capturing sentiment changes. On the visual modality of the CMU-MOSEI dataset, The MAE of the VAVL model is 0.795$\pm$0.028, which again is the lowest error rate compared to the baselines. The closest competitor is the MulT model with an MAE of 0.815$\pm$0.028, indicating that VAVL is more precise in interpreting visual data for sentiment analysis.

The MSP-IMPROV dataset further showcases the VAVL's robustness, particularly in the audio-visual category for arousal (Aro.), with a score of 0.856$\pm$0.110, significantly surpassing the UAVM model which has a score of 0.471$\pm$0.257. Similarly, in dominance (Dom.), VAVL scores 0.814$\pm$0.095, while the next best is the TSLTM model at 0.782$\pm$0.145. The results for the audio-visual setting on the MSP-IMPROV corpus reinforce our hypothesis that the use of averaging unimodal predictions, as employed by some baselines, might not work well for regression tasks. These metrics highlight the strong capability of the VAVL framework in detecting the intensity and control aspects of emotions conveyed through both audio and visual cues.

In summary, the VAVL model not only consistently outperforms the baseline models in terms of CCC and macro and micro F1 scores, it also maintains lower MAE across datasets, showcasing its superior ability to accurately recognize and predict emotions. The numerical superiority of the VAVL model across various datasets which have been collected under diverse scenarios emphasizes its potential for practical applications in emotion recognition systems.

\begin{figure}[tbp]
    \centering
    \includegraphics[trim={7.3cm 0 6cm 17.7cm},clip, width=0.7\columnwidth]{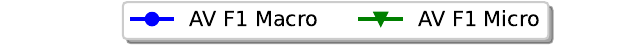}\\
    \subfigure[Audio Masking]
    {
        \includegraphics[trim={0.1cm 1.5cm 0.25cm 0.2cm},clip,width=4.2cm]{plots/plot_f1_scores_audio_av.pdf}
        \label{fig:f1_probs_scores_a}
    } 
    \subfigure[Video Masking]
    {
        \includegraphics[trim={0.1cm 1.5cm 0.25cm 0.2cm},clip,width=4.2cm]{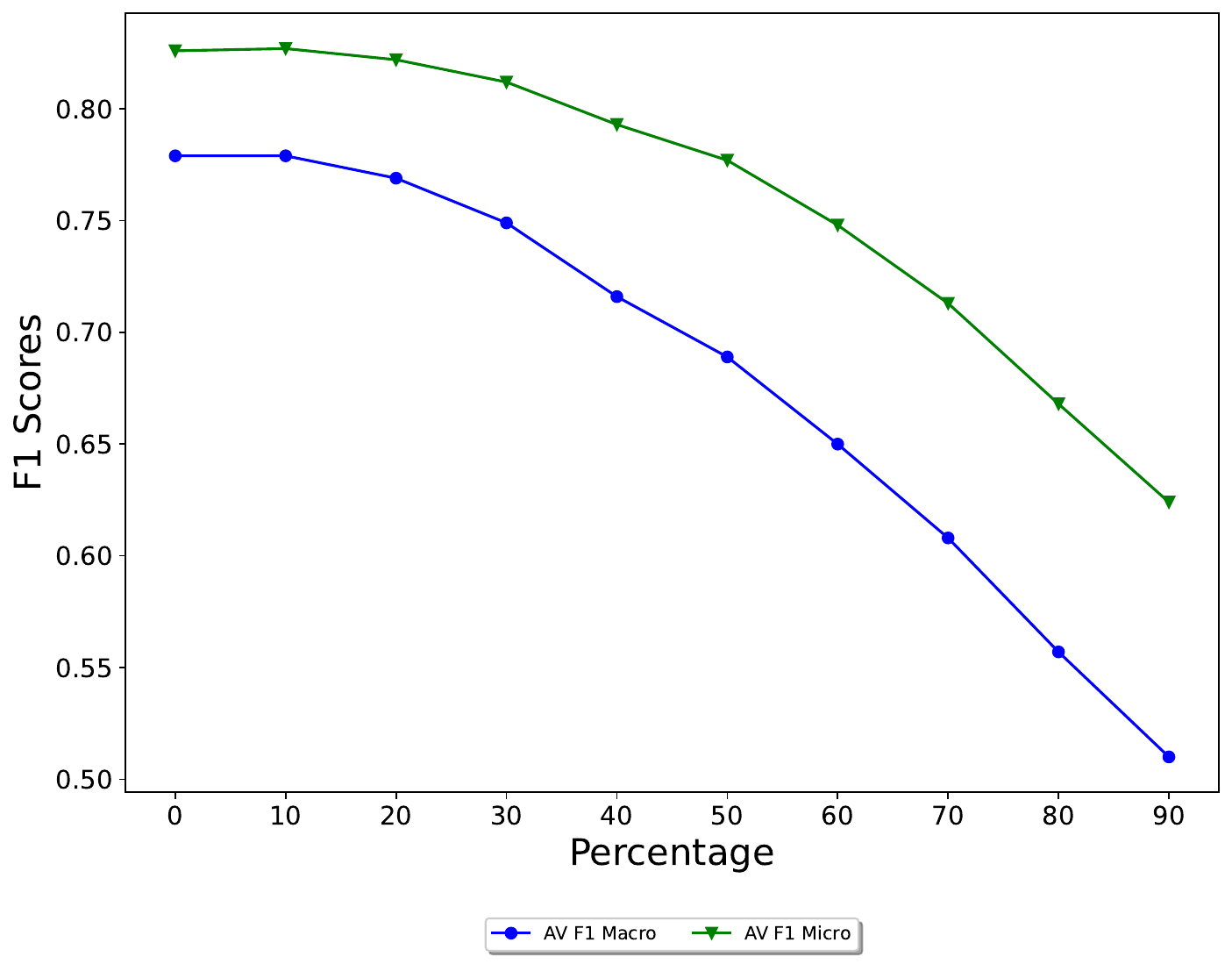}
        \label{fig:f1_probs_scores_b}
    }
\caption{Performance of the proposed model under audio-visual settings on the CREMA-D corpus with partial visual or acoustic information. The figure plots the micro F1-scores as a function of the percentage of the frames included for the masked modality (the other modality is assumed to be complete).}
\label{fig:probs_scores_crema}
\end{figure}

\begin{figure}[tbp]
    \centering
    \includegraphics[trim={7.3cm 0 6cm 17.7cm},clip, width=0.6\columnwidth]{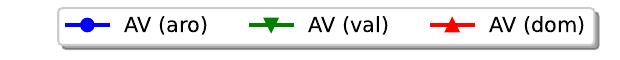}\\
    \subfigure[Audio Masking]
    {
        \includegraphics[trim={0.2cm 1cm 0cm 0.2cm},clip,width=4.2cm]{plots/plot_high_quality_audio_mask_av.pdf}
        \label{fig:ccc_probs_scores_a}
    } 
    \subfigure[Video Masking]
    {
        \includegraphics[trim={0.2cm 1cm 0cm 0.2cm},clip,width=4.2cm]{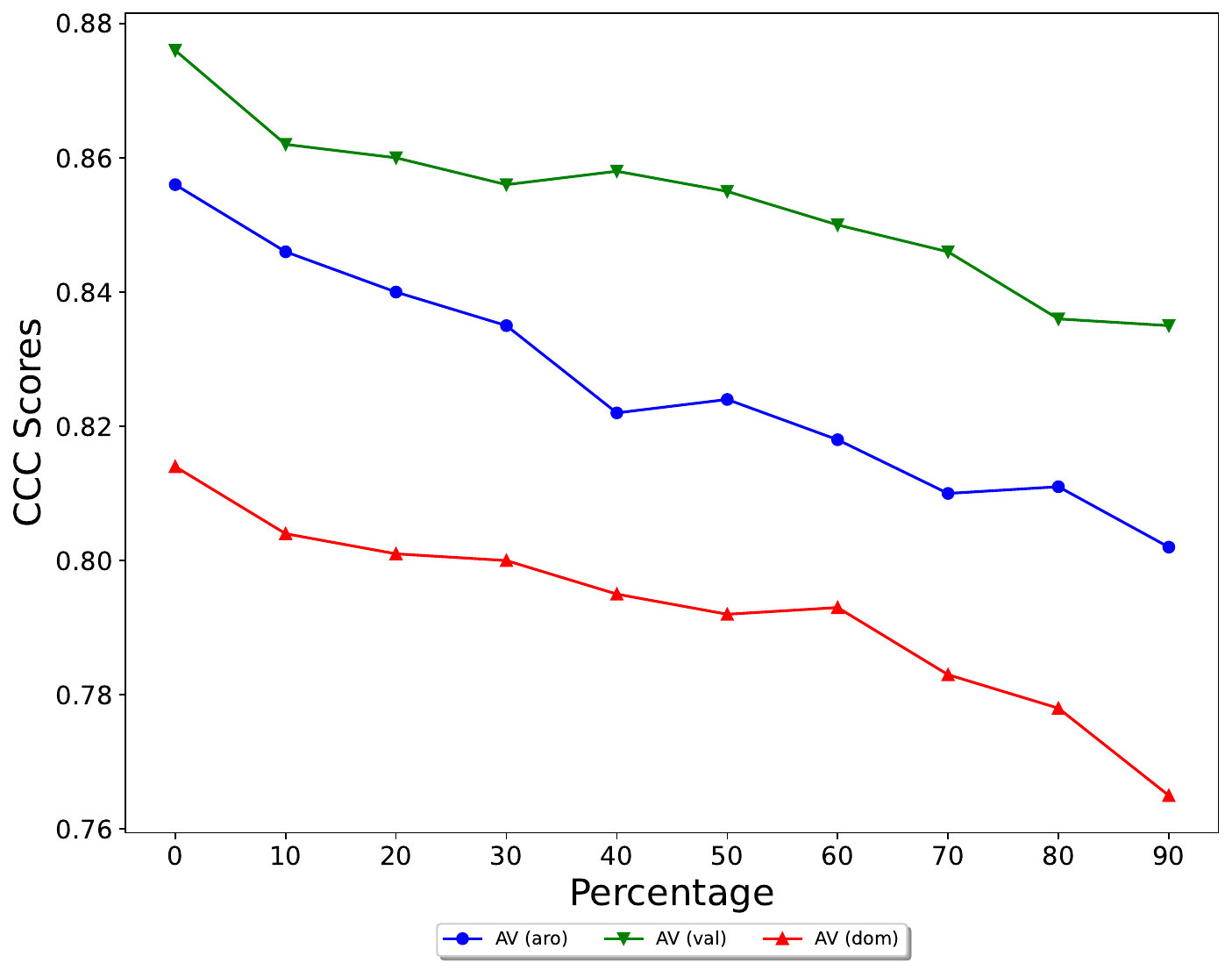}
        \label{fig:ccc_probs_scores_b}
    }
\caption{Performance of the proposed model  under audio-visual settings on the MSP-IMPROV corpus with partial visual or acoustic information. The figure plots the CCC scores as a function of the percentage of the frames included for the masked modality (the other modality is assumed to be complete).}
\label{fig:probs_scores_improv}
\end{figure}

\subsection{Random Masking of Features}
\label{sec:RandomMasking}

This section evaluates the system's performance in scenarios with absent features, simulating missing data by randomly zeroing out either visual or acoustic data at the frame level. To analyze the system's resilience to incomplete information, we incrementally mask available frames from 0\% to 90\% in 10\% increments. For instance, in the 30\% condition, 30\% of the frames from the modality being masked are replaced with zeros. This evaluation helps us understand the model's performance variations when audio-visual data is available but partially incomplete. The rationale behind our approach of simulating missing modalities by randomly dropping frames at varying percentages is to closely mimic conditions where input data might be incomplete or degraded due to practical issues. In real-world scenarios, data loss can occur randomly due to face occlusions, hardware intermittent malfunctions, or other environmental factors. Randomly dropping frames aims to replicate these unpredictable disruptions. We focus on the MSP-IMPROV and CREMA-D corpora, which allow us to use our entire framework—from processing raw inputs to predicting emotions - a process not possible with the CMU-MOSEI corpus due to the absence of raw data. The MSP-IMPROV and CREMA-D corpora provide consistency in frame-wise facial feature extraction and synchronize acoustic feature extraction timing, employing a 25 ms window size and a 20 ms stride for both corpora.

\begin{table*}[ht]
\captionsetup{justification=centering}
\caption{Ablation analysis of the proposed VAVL model using the MSP-IMPROV and CREMA-D corpora. Ablation 1 removes the residual connections over the shared layers. Ablation 2 removes reconstruction step from the framework. Ablation 3 removes audio-visual prediction layer, estimating the output by averaging the unimodal predictions.}
\label{tab:ablation}
\centering
\small
\setlength{\tabcolsep}{4pt} 
\begin{tabular*}{\textwidth}{@{\extracolsep{\fill}}l|*{9}{c}|*{6}{c}}
\toprule
& \multicolumn{9}{c|}{\textbf{MSP-IMPROV}} & \multicolumn{6}{c}{\textbf{CREMA-D}} \\
\cmidrule(lr){2-10} \cmidrule(lr){11-16}
 & \multicolumn{3}{c}{\textbf{Audio-Visual}} & \multicolumn{3}{c}{\textbf{Acoustic}} & \multicolumn{3}{c|}{\textbf{Visual}} & \multicolumn{2}{c}{\textbf{Audio-Visual}} & \multicolumn{2}{c}{\textbf{Acoustic}} & \multicolumn{2}{c}{\textbf{Visual}} \\
\cmidrule(lr){2-4} \cmidrule(lr){5-7} \cmidrule(lr){8-10} \cmidrule(lr){11-12} \cmidrule(lr){13-14} \cmidrule(lr){15-16} 
 \textbf{Model} & Aro. & Val. & Dom. & Aro. & Val. & Dom. & Aro. & Val. & Dom. & F1-Ma & F1-Mi & F1-Ma & F1-Mi & F1-Ma & F1-Mi \\
\midrule
VAVL & \textbf{0.856} & \textbf{0.876} & \textbf{0.814} & \textbf{0.853} & 0.858 & \textbf{0.783} & \textbf{0.422} & \textbf{0.631} & \textbf{0.375} & \textbf{0.779} & \textbf{0.826} & \textbf{0.628} & \textbf{0.701} & \textbf{0.738} & \textbf{0.787} \\
Ablt. 1 & 0.716 & 0.779 & 0.705 & 0.682 & 0.753 & 0.712 & 0.176 & 0.341 & 0.133 & 0.751 & 0.807 & 0.397 & 0.528 & 0.727 & 0.783 \\
Ablt. 2 & 0.810 & 0.873 & 0.788 & 0.791 & 0.864 & 0.780 & 0.172 & 0.577 & 0.227 & 0.761 & 0.816 & 0.604 & 0.684 & 0.718 & 0.775 \\
Ablt. 3 & 0.711 & 0.843 & 0.659 & 0.784 & \textbf{0.870} & 0.776 & 0.374 & 0.617 & 0.265 & 0.762 & 0.814 & 0.629 & 0.691 & 0.713 & 0.764 \\
\bottomrule
\end{tabular*}
\end{table*}

Figures \ref{fig:f1_probs_scores_a} and \ref{fig:f1_probs_scores_b} show the average results on the CREMA-D corpus for masking audio and visual inputs, respectively, obtained from the model's audio-visual prediction heads. The results indicate that removing either modality impacts the model. With random audio masking, performance remains stable when we mask up to 30\% of the frames. The model reaches an F1 Macro score of approximately 0.72 when only 10\% of the audio frames are present. In contrast, visual masking results in a sharper drop in performance when we mask 30\% of the frames. The approach has a low F1 Macro score of about 0.51 when only 10\% of the video frames are present. This score is significantly lower than those in Table 1 using only the acoustic head, implying that the audio-visual head may get confused with zero-masked frames. When the percentage of missing visual frames is higher than 60-70\%, it is better to rely solely on the acoustic modality.

Figures \ref{fig:ccc_probs_scores_a} and \ref{fig:ccc_probs_scores_b} depict average results on the MSP-IMPROV corpus for masking audio and visual inputs. The results from the audio-visual prediction heads show that the performance with random audio masking is steady up to 30\%, then drops sharply, reaching scores of about 0.4 CCC for valence, and 0.2 CCC  for arousal and dominance. This result suggests that the emotional attribute models rely more on the acoustic modality to achieve high performance. Masking the audio frames can lead to lower performances than using only the visual head for prediction. When we miss more than 60\% of the acoustic features, it is better to focus solely on visual features, using the results from the visual head. Conversely, masking visual features results in a smaller performance drop. Even when we mask 90\% of the visual features, we still achieve scores of 0.85 CCC for valence, 0.82 CCC for arousal, and 0.75 CCC for dominance. These findings confirm that the model is more dependent on acoustic features than visual features to predict emotional attributes.

\subsection{Ablation Analysis of the VAVL Framework}
\label{ssec:ablations}

Table \ref{tab:ablation} presents the results of an ablation analysis of the VAVL model on the MSP-IMPROV and CREMA-D corpora, comparing the performance of the full model with its ablated versions. In these experiments, we focus on the on the MSP-IMPROV and CREMA-D corpora, since with these corpora we were able to use our entire framework from raw inputs to emotion predictions, which was not possible with CMU-MOSEI since raw datapoints are not provided. Three ablated models are considered: Ablation 1 (Ablt. 1) removes the residual connections over the shared layers, Ablation 2 (Ablt. 2) removes the reconstruction step from the framework, and Ablation 3 (Ablt. 3) removes the audio-visual prediction layer, using the average of the unimodal predictions as the audio-visual prediction.

In the MSP-IMPROV dataset, the full VAVL model outperforms all the ablated versions in terms of arousal, valence, and dominance for the audio-visual and visual modalities. In the acoustic modality, VAVL shows the best performance for arousal and valence. However, Ablation 2 and 3 slightly surpass the VAVL results for dominance. These results indicate that the residual connections, reconstruction step, and audio-visual prediction layer all contribute to the strong performance of the VAVL model. 

On the CREMA-D dataset, VAVL consistently achieves the highest F1-Macro and F1-Micro scores across multimodal and unimodal settings, indicating that the full model is superior to its ablated versions. Ablation 1 has the lowest performance in the acoustic modality, while Ablations 2 and 3 show competitive results in some cases, albeit not surpassing the results of the full VAVL model. These results are consistent with the MSP-IMPROV ablation results and suggest that the combination of all components in the VAVL model leads to the best performance, with each component playing a significant role in the overall success of the model.

\subsection{Shared Embedding Analysis}
\label{ssec:embeddingAnalysis}
In this section, we perform analysis on the output embedding representations for each modality at the output of the shared layers from our VAVL model and the baselines. This experiment is conducted to investigate whether the shared layers of a multimodal model are capable of producing different representations for each modality. Using shared layers in a multimodal model allows the architecture to learn a common representation across different modalities, which can help in tasks that require integrating information from multiple sources. However, maintaining distinct representations for each modality in a multimodal model is crucial for several reasons. First, it allows the model to capture modality-specific information, which is essential for optimal performance. Second, it enhances interpretability by making it easier to analyze each modality's contribution to the final output or decision. Lastly, distinct representations ensure the model's adaptability, enabling it to perform well even when some modalities are missing or incomplete. To verify our proposed model's capability of generating distinct separable representations for each modality from the shared layers, we obtain separate output embeddings from the shared layers of the trained model using either acoustic or visual information. Then, we calculate the cosine distance between their respective embeddings. On the one hand, a high cosine distance indicates that the embeddings are more dissimilar, and thus capturing modality-specific information. On the other hand, a low cosine distance implies that the embeddings are more similar, potentially indicating that the model is not effectively capturing the unique features of each modality, and thus not effectively utilizing the multimodal input.  

\begin{figure}[t]
    \centering
    \includegraphics[clip,width=85mm,height=55mm]{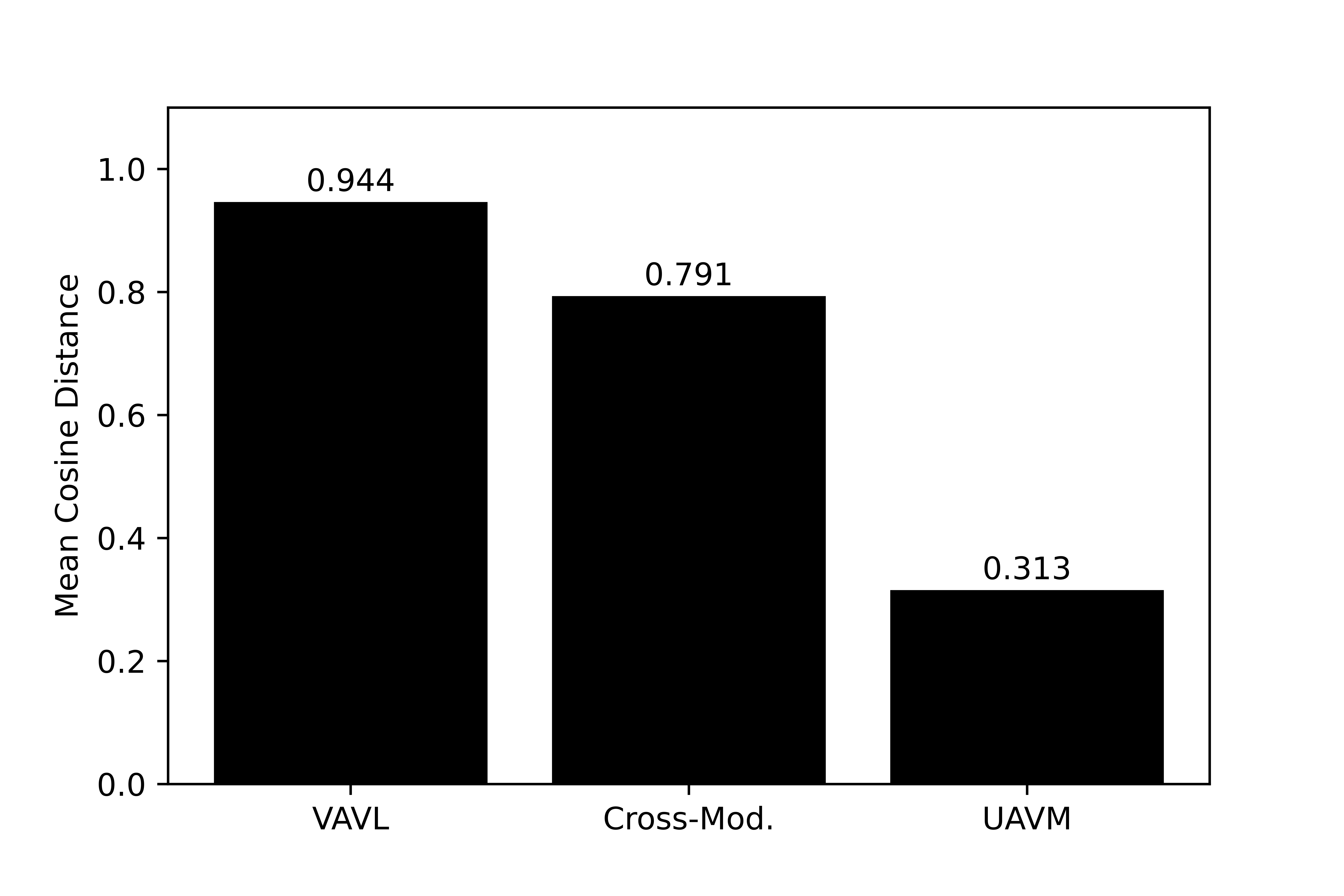}
    \caption{Comparison of average cosine distances between acoustic and visual representations generated by shared layers of the VAVL, AuxFormer, and UAVM models.}
    \label{fig:cosine}
\end{figure}

Figure \ref{fig:cosine} presents a comparison of the average cosine distances between acoustic and visual representations generated by the shared layers of our proposed framework (VAVL). For comparison, we use the shared models from the AuxFormer architecture \cite{Goncalves_2022}, which are implemented with cross-modal layers (the same type of layers used in the MulT model \cite{Tsai_2019}), and from the UAVM model \cite{gong_2022}. Based on the average cosine distance values obtained, the VAVL shared layers is the most effective model in generating distinct representations for the acoustic and visual modalities. The cross-modal layers show moderate effectiveness, while the UAVM shared layers seem to struggle in separating the modalities. These results correlate with our model's superior overall performance in unimodal settings compared to the baselines.

\section{Conclusions}
\label{sec:conclusion}

This study introduced the VAVL model, a novel versatile framework for audio-visual, audio-only, and video-only emotion recognition. The proposed approach is built with shared layers, residual connections over shared layers, and a unimodal reconstruction task. It attains high emotion recognition performance and can be trained even when paired audio and visual data is unavailable for part of the recordings. Furthermore, the model allows for direct application across both emotion regression and classification tasks. In the MSP-IMPROV corpus, the VAVL model consistently outperforms the CCC predictions of all multimodal baselines across all emotional attributes, achieving 0.856, 0.876, and 0.814 for arousal, valence, and dominance in audio-visual settings. Our results demonstrate that our method is able to better leverage visual representations to improve both unimodal and audio-visual performance, particularly in the visual-only setting. In the CREMA-D corpus, VAVL consistently surpasses baselines across all settings, achieving the highest F1-Macro (0.779) and F1-Micro (0.826) scores in the audio-visual modality. Under more naturalistic environment with the CMU-MOSEI corpus, we achieve better audio-visual and visual MAE scores than the strong baselines explored for sentiment score prediction. In addition to high performance on emotion recognition tasks, we show through the experimental evaluation our models' capabilities on handling missing modalities at the audio-visual setting and we also ran experiments to show that the shared layers of the proposed architecture are able to better capture the distinct features contained in acoustic and visual inputs, when compared to the shared layers of the baseline models.

While our training pipeline offers flexibility by accommodating both single and dual modalities, we acknowledge potential limitations. Specifically, training can become unstable when dealing with extremely highly imbalanced data from the two modalities. To mitigate this problem, we employ careful batch management and learning rate adjustments to ensure stability. Additionally, the iterative freezing and unfreezing of weights introduces some complexity to the training process, but these steps are crucial for effective learning from both unpaired and paired data.

Future work for this study is to utilize this framework on other audio-visual tasks such as sound recognition \cite{chen_2020_4}, scene classification \cite{wang_2021_3}, event localization \cite{tian_2018}, and audio-visual speech recognition \cite{Ma_2023_2}. Furthermore, extension to this work can include considering additional modalities, such as \emph{electroencephalogram} (EEG) \cite{Huang_2016_3}. Integrating EEG data with facial expressions could provide complementary information, enriching multi-modal emotion analysis and opening new research avenues. This approach is significantly less complex compared to other methods, such as cross-modal learning \cite{Tsai_2019, Goncalves_2022}. In cross-modal learning, bimodal context representation vectors are developed, and the introduction of each additional modality requires the addition of $M*(M-1)$ context layers to the model, where $M$ represents the number of modalities. We expect that our approach will be easier to computationally scale to incorporate new modalities.

\bibliographystyle{unsrt}  
\bibliography{references}

\begin{thebibliography}{10}

\bibitem{Schuller_2018}
B.~Schuller.
\newblock Speech emotion recognition: Two decades in a nutshell, benchmarks,
  and ongoing trends.
\newblock {\em Communications of the ACM}, 61(5):90--99, April 2018.

\bibitem{Tian_2001}
Y.-I. Tian, T.~Kanade, and J.~F. Cohn.
\newblock Recognizing action units for facial expression analysis.
\newblock {\em IEEE Transactions on Pattern Analysis and Machine Intelligence},
  23(2):97--115, February 2001.

\bibitem{Mariooryad_2016}
S.~Mariooryad and C.~Busso.
\newblock Facial expression recognition in the presence of speech using blind
  lexical compensation.
\newblock {\em IEEE Transactions on Affective Computing}, 7(4):346--359,
  October-December 2016.

\bibitem{Busso_2006}
C.~Busso and S.S. Narayanan.
\newblock Interplay between linguistic and affective goals in facial expression
  during emotional utterances.
\newblock In {\em 7th International Seminar on Speech Production (ISSP 2006)},
  pages 549--556, Ubatuba-SP, Brazil, December 2006.

\bibitem{Busso_2004}
C.~Busso, Z.~Deng, S.~Yildirim, M.~Bulut, C.M. Lee, A.~Kazemzadeh, S.~Lee,
  U.~Neumann, and S.~Narayanan.
\newblock Analysis of emotion recognition using facial expressions, speech and
  multimodal information.
\newblock In {\em Sixth International Conference on Multimodal Interfaces ICMI
  2004}, pages 205--211, State College, PA, October 2004. ACM Press.

\bibitem{DMello_2015}
S.~K. D'mello and J.~Kory.
\newblock A review and meta-analysis of multimodal affect detection systems.
\newblock {\em ACM Computing Surveys}, 47(3):1--36, April 2015.

\bibitem{Khare_2021}
A.~Khare, S.~Parthasarathy, and S.~Sundaram.
\newblock Self-supervised learning with cross-modal transformers for emotion
  recognition.
\newblock In {\em IEEE Spoken Language Technology Workshop (SLT 2021)}, pages
  381--388, Shenzhen, China, January 2021.

\bibitem{Bouali_2022}
Y.~L. Bouali, O.~B. Ahmed, and S.~Mazouzi.
\newblock Cross-modal learning for audio-visual emotion recognition in acted
  speech.
\newblock In {\em International Conference on Advanced Technologies for Signal
  and Image Processing (ATSIP 2022)}, pages 1--6, Sfax, Tunisia, May 2022.

\bibitem{Tran_2022}
M.~Tran and M.~Soleymani.
\newblock A pre-trained audio-visual transformer for emotion recognition.
\newblock In {\em IEEE International Conference on Acoustics, Speech and Signal
  Processing (ICASSP 2022)}, pages 4698--4702, Singapore, Singapore, May 2022.

\bibitem{Zhu_2021_2}
H.~Zhu, M.-D. Luo, R.~Wang, A.-H. Zheng, and R.~He.
\newblock Deep audio-visual learning: A survey.
\newblock {\em International Journal of Automation and Computing}, 18:351--376,
  June 2021.

\bibitem{Parthasarathy_2020_2}
S.~Parthasarathy and S.~Sundaram.
\newblock Training strategies to handle missing modalities for audio-visual
  expression recognition.
\newblock In {\em International Conference on Multimodal Interaction (ICMI
  2020)}, pages 400--404, Utrecht, The Netherlands, October 2020.

\bibitem{Goncalves_2022_3}
L.~Goncalves and C.~Busso.
\newblock Robust audiovisual emotion recognition: Aligning modalities,
  capturing temporal information, and handling missing features.
\newblock {\em IEEE Transactions on Affective Computing}, 13(4):2156--2170,
  October-December 2022.

\bibitem{Ngiam_2011}
J.~Ngiam, A.~Khosla, M.~Kim, J.~Nam, H.~Lee, and A.~Ng.
\newblock Multimodal deep learning.
\newblock In {\em International conference on machine learning (ICML2011)},
  pages 689--696, Bellevue, WA, USA, June-July 2011.

\bibitem{arandjelovic_2017}
R.~Arandjelovi\'{c} and A.~Zisserman.
\newblock Look, listen and learn.
\newblock In {\em IEEE International Conference on Computer Vision (ICCV
  2017)}, pages 609--617, Venice, Italy, October 2017.

\bibitem{Antoniadis_2021}
P.~Antoniadis, I.~Pikoulis, P.P. Filntisis, and P.~Maragos.
\newblock An audiovisual and contextual approach for categorical and continuous
  emotion recognition in-the-wild.
\newblock In {\em IEEE/CVF International Conference on Computer Vision
  Workshops (ICCVW 2021)}, pages 3638--3644, Montreal, BC, Canada, October
  2021.

\bibitem{Hao_2020}
M.~Hao, W.-H. Cao, Z.-T. Liu, M.~Wu, and P.~Xiao.
\newblock Visual-audio emotion recognition based on multi-task and ensemble
  learning with multiple features.
\newblock {\em Neurocomputing}, 391:42--51, May 2020.

\bibitem{Vaswani_2017}
A.~Vaswani, N.~Shazeer, N.~Parmar, J.~Uszkoreit, L.~Jones, A.N. Gomez,
  {\L}.~Kaiser, and I.~Polosukhin.
\newblock Attention is all you need.
\newblock In {\em In Advances in Neural Information Processing Systems (NIPS
  2017)}, pages 5998--6008, Long Beach, CA, USA, December 2017.

\bibitem{Akbari_2021}
H.~Akbari, L.~Yuan, R.~Qian, W.-H. Chuang, S.-F. Chang, Y.~Cui, and B.~Gong.
\newblock {VATT}: Transformers for multimodal self-supervised learning from raw
  video, audio and text.
\newblock In {\em Conference on Neural Information Processing Systems (NeurIPS
  2021)}, pages 24206--24221, Virtual Conference, December 2021.

\bibitem{gong_2022}
Y.~Gong, A.~H. Liu, A.~Rouditchenko, and J.~Glass.
\newblock {UAVM}: Towards unifying audio and visual models.
\newblock {\em IEEE Signal Processing Letters}, 29:2437--2441, November 2022.

\bibitem{Baltrusaitis_2019}
T.~Baltru{\v{s}}aitis, C.~Ahuja, and L.~P. Morency.
\newblock Multimodal machine learning: A survey and taxonomy.
\newblock {\em IEEE Transactions on Pattern Analysis and Machine Intelligence},
  41(2):423--443, February 2019.

\bibitem{ElAyadi_2011}
M.~{El Ayadi}, M.S. Kamel, and F.~Karray.
\newblock Survey on speech emotion recognition: Features, classification
  schemes, and databases.
\newblock {\em Pattern Recognition}, 44(3):572--587, March 2011.

\bibitem{Zadeh_2018_2}
A.~Zadeh, P.P. Liang, S.~Poria, P.~Vij, E.~Cambria, and L.-P. Morency.
\newblock Multi-attention recurrent network for human communication
  comprehension.
\newblock In {\em AAAI Conference on Artificial Intelligence (AAAI 2018)},
  volume~32, pages 5642--5649, New Orleans, LA, USA, February 2018.

\bibitem{agrawal_2016}
A.~Agrawal, D.~Batra, and D.~Parikh.
\newblock Analyzing the behavior of visual question answering models.
\newblock In {\em Conference on Empirical Methods in Natural Language
  Processing (EMNLP 2016)}, pages 1955--1960, Austin, TX, USA, November 2016.

\bibitem{Liang_2022}
P.P. Liang, A.~Zadeh, and L.-P. Morency.
\newblock Foundations and trends in multimodal machine learning: Principles,
  challenges, and open questions.
\newblock {\em ArXiv e-prints (arXiv:2209.03430)}, pages 1--36, September 2022.

\bibitem{Cao_2014_2}
H.~Cao, D.G. Cooper, M.K. Keutmann, R.C. Gur, A.~Nenkova, and R.~Verma.
\newblock {CREMA-D}: Crowd-sourced emotional multimodal actors dataset.
\newblock {\em IEEE Transactions on Affective Computing}, 5(4):377--390,
  October-December 2014.

\bibitem{Busso_2017}
C.~Busso, S.~Parthasarathy, A.~Burmania, M.~AbdelWahab, N.~Sadoughi, and
  E.~{Mower Provost}.
\newblock {MSP-IMPROV}: An acted corpus of dyadic interactions to study emotion
  perception.
\newblock {\em IEEE Transactions on Affective Computing}, 8(1):67--80,
  January-March 2017.

\bibitem{Parthasarathy_2017}
S.~Parthasarathy, R.~Lotfian, and C.~Busso.
\newblock Ranking emotional attributes with deep neural networks.
\newblock In {\em IEEE International Conference on Acoustics, Speech and Signal
  Processing (ICASSP 2017)}, pages 4995--4999, New Orleans, LA, USA, March
  2017.

\bibitem{Lotfian_2018}
R.~Lotfian and C.~Busso.
\newblock Predicting categorical emotions by jointly learning primary and
  secondary emotions through multitask learning.
\newblock In {\em Interspeech 2018}, pages 951--955, Hyderabad, India,
  September 2018.

\bibitem{savchenko_2022}
A.~V. Savchenko, L.~V. Savchenko, and I.~Makarov.
\newblock Classifying emotions and engagement in online learning based on a
  single facial expression recognition neural network.
\newblock {\em IEEE Transactions on Affective Computing}, 13(4):2132--2143,
  October-December 2022.

\bibitem{Lin_2023_3}
W.-C. Lin, L.~Goncalves, and C.~Busso.
\newblock Enhancing resilience to missing data in audio-text emotion
  recognition with multi-scale chunk regularization.
\newblock In {\em ACM International Conference on Multimodal Interaction (ICMI
  2023)}, pages 207--215, Paris, France, October 2023.

\bibitem{Li_2023}
J.~Li, Y.~Chen, X.~Zhang, J.~Nie, Z.~Li, Y.~Yu, Y.~Zhang, R.~Hong, and M.~Wang.
\newblock Multimodal feature extraction and fusion for emotional reaction
  intensity estimation and expression classification in videos with
  transformers.
\newblock {\em ArXiv e-prints (arXiv:2303.09164)}, pages 1--7, March 2023.

\bibitem{Yu_2023}
J.~Yu, J.~Zhu, W.~Zhu, Z.~Cai, G.~Xie, R.~Li, and G.~Zhao.
\newblock A dual branch network for emotional reaction intensity estimation.
\newblock {\em ArXiv e-prints (arXiv:2303.09210)}, pages 1--6, March 2023.

\bibitem{Goncalves_2023}
L.~Goncalves and C.~Busso.
\newblock Learning cross-modal audiovisual representations with ladder networks
  for emotion recognition.
\newblock In {\em IEEE International Conference on Acoustics, Speech and Signal
  Processing (ICASSP 2023)}, pages 1--5, Rhodes island, Greece, June 2023.

\bibitem{Ekman_1971}
P.~Ekman and W.V. Friesen.
\newblock Constants across cultures in the face and emotion.
\newblock {\em Journal of Personality and Social Psychology}, 17(2):124--129,
  March 1971.

\bibitem{Goncalves_2022_2}
L.~Goncalves and C.~Busso.
\newblock Improving speech emotion recognition using self-supervised learning
  with domain-specific audiovisual tasks.
\newblock In {\em Interspeech 2022}, pages 1168--1172, Incheon, South Korea,
  September 2022.

\bibitem{ghaleb_2020}
E.~Ghaleb, M.~Popa, and S.~Asteriadis.
\newblock Metric learning-based multimodal audio-visual emotion recognition.
\newblock {\em IEEE MultiMedia}, 27(1):37--48, January-March 2020.

\bibitem{Mollahosseini_2019}
A.~Mollahosseini, B.~Hasani, and M.~H. Mahoor.
\newblock {AffectNet}: A database for facial expression, valence, and arousal
  computing in the wild.
\newblock {\em IEEE Transactions on Affective Computing}, 10(1):18--31,
  January-March 2019.

\bibitem{Dhall_2012}
A.~Dhall, R.~Goecke, S.~Lucey, and T.~Gedeon.
\newblock Collecting large, richly annotated facial-expression databases from
  movies.
\newblock {\em IEEE Multimedia}, 19(3):34--41, July-September 2012.

\bibitem{Bradley_1994}
M.M. Bradley and P.J. Lang.
\newblock Measuring emotion: the self-assessment manikin and the semantic
  differential.
\newblock {\em Journal of Behavior Therapy and Experimental Psychiatry},
  25(1):49--59, March 1994.

\bibitem{Busso_2008_5}
C.~Busso, M.~Bulut, C.C. Lee, A.~Kazemzadeh, E.~Mower, S.~Kim, J.N. Chang,
  S.~Lee, and S.S. Narayanan.
\newblock {IEMOCAP}: Interactive emotional dyadic motion capture database.
\newblock {\em Journal of Language Resources and Evaluation}, 42(4):335--359,
  December 2008.

\bibitem{Schoneveld_2021}
L.~Schoneveld, A.~Othmani, and H.~Abdelkawy.
\newblock Leveraging recent advances in deep learning for audio-visual emotion
  recognition.
\newblock {\em Pattern Recognition Letters}, 146:1--7, June 2021.

\bibitem{atmaja_2020}
B.~T. Atmaja and M.~Akagi.
\newblock Multitask learning and multistage fusion for dimensional audiovisual
  emotion recognition.
\newblock In {\em IEEE International Conference on Acoustics, Speech and Signal
  Processing (ICASSP 2020)}, pages 4482--4486, Barcelona, Spain, May 2020.

\bibitem{Hsu_2023}
J.-H. Hsu and C.~H. Wu.
\newblock Applying segment-level attention on bi-modal transformer encoder for
  audio-visual emotion recognition.
\newblock {\em IEEE Transactions on Affective Computing}, 14(4):3231--3243,
  October-December 2023.

\bibitem{wasi_2023}
A.T. Wasi, K.~\v{S}erbetar, R.~Islam, T.H. Rafi, and D.-K. Chae.
\newblock {ARBEx}: Attentive feature extraction with reliability balancing for
  robust facial expression learning.
\newblock {\em ArXiv e-prints (arXiv:2305.01486)}, pages 1--12, May 2023.

\bibitem{Wagner_2022}
J.~Wagner, A.~Triantafyllopoulos, H.~Wierstorf, M.~Schmitt, F.~Burkhardt,
  F.~Eyben, and B.W. Schuller.
\newblock Dawn of the transformer era in speech emotion recognition: closing
  the valence gap.
\newblock {\em ArXiv e-prints (arXiv:2203.07378)}, pages 1--25, March 2022.

\bibitem{Upadhyay_2023_2}
S.G. Upadhyay, W.-S. Chien, B.-H. Su, L.~Goncalves, Y.-T. Wu, A.N. Salman,
  C.~Busso, and C.-C. Lee.
\newblock An intelligent infrastructure toward large scale naturalistic
  affective speech corpora collection.
\newblock In {\em International Conference on Affective Computing and
  Intelligent Interaction (ACII 2023)}, pages 1--8, Cambridge, MA, USA,
  September 2023.

\bibitem{jaegle_2021}
A.~Jaegle, F.~Gimeno, A.~Brock, O.~Vinyals, A.~Zisserman, and J.~Carreira.
\newblock Perceiver: General perception with iterative attention.
\newblock In M.~Meila, , and T.~Zhang, editors, {\em International Conference
  on Machine Learning (ICML 2021)}, volume 139, pages 4651--4664. Proceedings
  of Machine Learning Research (PMLR), Virtual, July 2021.

\bibitem{baevski_2022}
A.~Baevski, W.-N. Hsu, Q.~Xu, A.~Babu, J.~Gu, and M.~Auli.
\newblock Data2vec: A general framework for self-supervised learning in speech,
  vision and language.
\newblock In K.~Chaudhuri, S.~Jegelka, L.~Song, C.~Szepesvari, G.~Niu, and
  S.~Sabato, editors, {\em International Conference on Machine Learning (ICML
  2022)}, volume 162, pages 1298--1312. Proceedings of Machine Learning
  Research (PMLR), Honolulu, HI, USA, July 2022.

\bibitem{dai_2022}
Y.~Dai, Duyu Tang, Liangxin Liu, Minghuan Tan, Cong Zhou, Jingquan Wang,
  Zhangyin Feng, Fan Zhang, Xueyu Hu, and Shuming Shi.
\newblock One model, multiple modalities: A sparsely activated approach for
  text, sound, image, video and code.
\newblock {\em ArXiv e-prints (arXiv:2205.06126)}, pages 1--14, May 2022.

\bibitem{shvetsova_2022}
N.~Shvetsova, B.~Chen, A.~Rouditchenko, S.~Thomas, B.~Kingsbury, R.~Feris,
  D.~Harwath, J.~Glass, and H.~Kuehne.
\newblock Everything at once-multi-modal fusion transformer for video
  retrieval.
\newblock In {\em IEEE/CVF Conference on Computer Vision and Pattern
  Recognition (CVPR 2022)}, pages 19988--19997, New Orleans, LA, USA, June
  2022.

\bibitem{Goncalves_2022}
L.~Goncalves and C.~Busso.
\newblock {AuxFormer}: Robust approach to audiovisual emotion recognition.
\newblock In {\em IEEE International Conference on Acoustics, Speech and Signal
  Processing (ICASSP 2022)}, pages 7357--7361, Singapore, May 2022.

\bibitem{Guo_2018}
X.~Guo, B.~Zhu, L.F. Polan\'{i}a, C.~Boncelet, and K.E. Barner.
\newblock Group-level emotion recognition using hybrid deep models based on
  faces, scenes, skeletons and visual attentions.
\newblock In {\em ACM International Conference on Multimodal Interaction (ICMI
  2018)}, pages 635--639, Boulder, CO, USA, October 2018.

\bibitem{Khan_2018}
A.S. Khan, Z.~Li, J.~Cai, Z.~Meng, J.~O'Reilly, and Y.~Tong.
\newblock Group-level emotion recognition using deep models with a four-stream
  hybrid network.
\newblock In {\em ACM International Conference on Multimodal Interaction (ICMI
  2018)}, pages 623--629, Boulder, CO, USA, October 2018.

\bibitem{gulati_2020}
A.~Gulati, J.~Qin, C.-C. Chiu, N.~Parmar, Y.~Zhang, J.~Yu, W.~Han, S.~Wang,
  Z.~Zhang, Y.~Wu, and R.~Pang.
\newblock Conformer: Convolution-augmented transformer for speech recognition.
\newblock In {\em Interspeech 2020}, pages 5036--5040, Shanghai, China, October
  2020.

\bibitem{Hsu_2021_2}
W.-N. Hsu, A.~Sriram, A.~Baevski, T.~Likhomanenko, Q.~Xu, V.~Pratap, J.~Kahn,
  A.~Lee, R.~Collobert, G.~Synnaeve, and M.~Auli.
\newblock Robust {wav2vec 2.0}: Analyzing domain shift in self-supervised
  pre-training.
\newblock {\em ArXiv e-prints (arXiv:2104.01027)}, pages 1--9, April 2021.

\bibitem{Baevski_2020}
A.~Baevski, Y.~Zhou, A.~Mohamed, and M.~Auli.
\newblock wav2vec 2.0: A framework for self-supervised learning of speech
  representations.
\newblock In {\em Advances in Neural Information Processing Systems (NeurIPS
  2020)}, volume~33, pages 12449--12460, Virtual, December 2020.

\bibitem{Wolf_2019}
T.~Wolf, L.~Debut, V.~Sanh, J.~Chaumond, C.~Delangue, A.~Moi, P.~Cistac,
  T.~Rault, R.~Louf, M.~Funtowicz, J.~Davison, S.~Shleifer, P.~{von Platen},
  C.~Ma, Y.~Jernite, J.~Plu, C.~Xu, T.~{Le Scao}, S.~Gugger, M.~Drame, and
  Q.~Lhoest amd A.M.~Rush.
\newblock {HuggingFace's} transformers: State-of-the-art natural language
  processing.
\newblock {\em ArXiv e-prints (arXiv:1910.03771v5)}, pages 1--8, October 2019.

\bibitem{Lotfian_2019_3}
R.~Lotfian and C.~Busso.
\newblock Building naturalistic emotionally balanced speech corpus by
  retrieving emotional speech from existing podcast recordings.
\newblock {\em IEEE Transactions on Affective Computing}, 10(4):471--483,
  October-December 2019.

\bibitem{Kingma_2014_2}
D.P. Kingma and J.~Ba.
\newblock Adam: A method for stochastic optimization.
\newblock In {\em International Conference on Learning Representations}, pages
  1--13, San Diego, CA, USA, May 2015.

\bibitem{Zhang_2016_8}
K.~Zhang, Z.~Zhang, Z.~Li, and Y.~Qiao.
\newblock Joint face detection and alignment using multitask cascaded
  convolutional networks.
\newblock {\em IEEE Signal Processing Letters}, 23(10):1499--1503, October
  2016.

\bibitem{Tsai_2019}
Y.-H.H. Tsai, S.~Bai, P.P. Liang, J.Z. Kolter, L.-P. Morency, and
  R.~Salakhutdinov.
\newblock Multimodal transformer for unaligned multimodal language sequences.
\newblock In {\em Association for Computational Linguistics (ACL 2019)},
  volume~1, pages 6558--6569, Florence, Italy, July 2019.

\bibitem{Chumachenko_2022}
K.~Chumachenko, A.~Iosifidis, and M.~Gabbouj.
\newblock Self-attention fusion for audiovisual emotion recognition with
  incomplete data.
\newblock In {\em IEEE International Conference on Pattern Recognition (ICPR
  2022)}, pages 2822--2828, Montreal, QC, Canada, August 2022.

\bibitem{Huang_2020}
J.~Huang, J.~Tao, B.~Liu, Z.~Lian, and M.~Niu.
\newblock Multimodal transformer fusion for continuous emotion recognition.
\newblock In {\em IEEE International Conference on Acoustics, Speech and Signal
  Processing (ICASSP 2020)}, pages 3507--3511, Barcelona, Spain, May 2020.

\bibitem{chen_2020_4}
H.~Chen, W.~Xie, A.~Vedaldi, and A.~Zisserman.
\newblock Vggsound: A large-scale audio-visual dataset.
\newblock In {\em IEEE International Conference on Acoustics, Speech and Signal
  Processing (ICASSP 2020)}, pages 721--725, Barcelona, Spain, May 2020.

\bibitem{wang_2021_3}
S.~Wang, A.~Mesaros, T.~Heittola, and T.~Virtanen.
\newblock A curated dataset of urban scenes for audio-visual scene analysis.
\newblock In {\em IEEE International Conference on Acoustics, Speech and Signal
  Processing (ICASSP 2021)}, pages 626--630, Toronto, ON, Canada, June 2021.

\bibitem{tian_2018}
Y.~Tian, J.~Shi, B.~Li, Z.~Duan, and C.~Xu.
\newblock Audio-visual event localization in unconstrained videos.
\newblock In V.~Ferrari, M.~Hebert, C.~Sminchisescu, and Y.~Weiss, editors,
  {\em European Conference on Computer Vision (ECCV 2018)}, volume 11206 of
  {\em Lecture Notes in Computer Science}, pages 252--268. Springer Berlin
  Heidelberg, Munich, Germany, September 2018.

\bibitem{Ma_2023_2}
P.~Ma, A.~Haliassos, A.~Fernandez-Lopez, H.~Chen, S.~Petridis, and M.~Pantic.
\newblock {Auto-AVSR}: Audio-visual speech recognition with automatic labels.
\newblock In {\em IEEE International Conference on Acoustics, Speech and Signal
  Processing (ICASSP 2023)}, pages 1--5, Rhodes Island, Greece, June 2023.

\bibitem{Huang_2016_3}
X.~Huang, J.~Kortelainen, G.~Zhao, X.~Li, A.~Moilanen, T.~Sepp\"{a}nen, and
  M.~Pietik\"{a}inen.
\newblock Multi-modal emotion analysis from facial expressions and
  electroencephalogram.
\newblock {\em Computer Vision and Image Understanding}, 147:114--124, June
  2016.

\end{thebibliography}

\end{document}